\documentclass{article}
\usepackage{microtype}
\usepackage{graphicx}
\usepackage{subcaption}
\usepackage{booktabs} 
\usepackage{float}
\usepackage{placeins}
\usepackage{hyperref}

\usepackage[mathscr]{eucal}
\usepackage{enumitem}

\usepackage[accepted]{preprint}

\usepackage{amsmath}
\usepackage{amssymb}
\usepackage{mathtools}
\usepackage{amsthm}

\theoremstyle{plain}
\newtheorem{theorem}{Theorem}[section]

\theoremstyle{remark}

\usepackage{multirow}

\usepackage{amsmath,amsfonts,bm}

\def\1{\bm{1}}

\DeclareMathAlphabet{\mathsfit}{\encodingdefault}{\sfdefault}{m}{sl}
\SetMathAlphabet{\mathsfit}{bold}{\encodingdefault}{\sfdefault}{bx}{n}

\newcommand{\cali}{\mathcal}

\newcommand{\incc}{\subseteq}

\newcommand{\E}{\mathbb{E}}

\newcommand{\R}{\mathbb{R}}

\usepackage{tikz}
\usetikzlibrary{positioning}
\usetikzlibrary{shapes.geometric}
\usetikzlibrary{shapes,snakes}
\usetikzlibrary{arrows.meta}
\usetikzlibrary{arrows, positioning, quotes}
\tikzstyle{arrow} = [draw, -latex']

\usepackage{wrapfig, framed, caption}
\usepackage{multirow}
\hypersetup{colorlinks,linkcolor={blue},citecolor={blue},urlcolor={black}}  

\usepackage[textsize=tiny]{todonotes}

\usepackage[capitalize,noabbrev]{cleveref}

\usepackage[textsize=tiny]{todonotes}

\icmltitlerunning{Neural Polar Factorization}
\usepackage{pythonhighlight}
\begin{document}
\include{math_commands}
\twocolumn[
\icmltitle{On a Neural Implementation of Brenier's Polar Factorization}

\begin{icmlauthorlist}
\icmlauthor{Nina Vesseron}{ensae}
\icmlauthor{Marco Cuturi}{app,ensae}
\end{icmlauthorlist}
\icmlaffiliation{app}{Apple}
\icmlaffiliation{ensae}{CREST-ENSAE, IP Paris}

\icmlkeywords{Machine Learning, ICML}

\vskip 0.3in
]

\printAffiliationsAndNotice{}  %

\begin{abstract}
In ~\citeyear{Brenier1991PolarFA}, \citeauthor{Brenier1991PolarFA} proved a theorem that generalizes the polar decomposition for square matrices -- factored as PSD $\times$ unitary -- to any vector field $F:\mathbb{R}^d\rightarrow \mathbb{R}^d$. The theorem, known as the polar factorization theorem, states that any field $F$ can be recovered as the composition of the gradient of a convex function $u$ with a measure-preserving map $M$, namely $F=\nabla u \circ M$. We propose a practical implementation of this far-reaching theoretical result, and explore possible uses within machine learning. The theorem is closely related to optimal transport (OT) theory, and we borrow from recent advances in the field of neural optimal transport to parameterize the potential $u$ as an input convex neural network. The map $M$ can be either evaluated pointwise using $u^*$, the convex conjugate of $u$, through the identity $M=\nabla u^* \circ F$, or learned as an auxiliary network. Because $M$ is, in general, not injective, we consider the additional task of estimating the ill-posed inverse map that can approximate the pre-image measure $M^{-1}$ using a stochastic generator. We illustrate possible applications of \citeauthor{Brenier1991PolarFA}'s polar factorization to non-convex optimization problems, as well as sampling of densities that are not log-concave.
\end{abstract}

\section{Introduction}
\citeauthor{Brenier1991PolarFA} proved, through his seminal polar factorization theorem~\citeyearpar{Brenier1991PolarFA}, that \textit{any} vector field can be decomposed into two simpler elements: Given a reference measure $\rho$ supported on $\Omega\subset\mathbb{R}^d$, for any $F:\Omega\rightarrow \mathbb{R}^d$, there exists a convex potential $u:\mathbb{R}^d\rightarrow \mathbb{R}$ and a \textit{measure-preserving} map $M:\Omega\rightarrow \Omega$ (i.e. one has $M_\sharp\rho=\rho$), such that $F=\nabla u \circ M$. 
The polar factorization theorem states that \textit{any} vector field, no matter how irregular, can be reshuffled to match that of the gradient of a convex potential, and that this careful reshuffling in space is achieved by the measure-preserving map $M$. This paper aims to provide a practical approach to recover approximations of the potential $u$ and vector-valued map $M$ using exclusively samples $x_i\sim\rho$ and their associated images $F(x_i)$. We also highlight how a reliable polar factorization solver, coupled with an estimation of a stochastic generator that mimics the measure-valued inverse map $M^{-1}$, can be used to study the gradient field of non-convex landscapes. We consider, in particular, the case where the field $F$ of interest is the gradient, with respect to the parameters of a neural architecture, of a learning loss.
Note that the polar factorization theorem should not be confused with the major theorem from optimal transport closely associated  with~\citeauthor{Brenier1991PolarFA}, which we recall in \S\ref{sec:back}. That theorem states that the~\citeauthor{monge1781memoire} formulation of the optimal transport (OT) problem, which seeks the push-forward map transporting a measure onto another with the least mean displacements (as measured with squared norms) is solved by the gradient of a convex potential. 

\paragraph{Existing Implementations.} Shortly after~\citep{Brenier1991PolarFA}, ~\citet{benamou1994domain} proposed a numerical approach to decompose a vector field, with an explicit Eulerian (gridded) approach. Lagrangian approaches have been proposed by~\citet{Gallout2018ALS}, while~\citet{doi:10.1137/15M1017235} use a semidiscrete OT formulation. Both are used on low-dimensional manifolds as lower level subroutines to solve Euler’s equation for incompressible and
inviscid fluids~\citep{arnold1966geometrie}. More recently, \citet{morel2023turning} proposed to use \citeauthor{Brenier1991PolarFA}'s insight to gradually refactor a normalizing flow as the gradient of a convex map, a.k.a a \citet{monge1781memoire} map, by applying measure-preserving maps for the Gaussian distribution. Their approach does not, however, rely on neural OT solvers, and focuses instead on untangling an existing flow to turn it gradually into the gradient of a convex potential.

\paragraph{Contributions.} We propose in this work a neural implementation of the polar factorization theorem that leverages recent advances in neural optimal transport. More precisely,
\begin{itemize}[leftmargin=.3cm,itemsep=.0cm,topsep=0cm,parsep=2pt]
    \item After introducing the polar factorization theorem, as well as neural OT solvers, we show how the two blocks of \citeauthor{Brenier1991PolarFA}'s result can be recovered using input convex neural networks (ICNN)~\citep{pmlr-v70-amos17b}. We modify the ICNN architecture originally proposed in~\citet{pmlr-v70-amos17b} and ~\citet{korotin2020wasserstein2}, to propose quadratic (low-rank + diagonal) positive definite layers at each layer. Starting from an arbitrary field $F$, we use the modifications proposed by~\citet{amos2023amortizing} to train the~\citeauthor{Brenier1991PolarFA} convex potential $u_\theta$, that appears in the polar factorization of $F$.
    \item We study two alternative parameterizations for the measure-preserving map $M$: Either implicit, relying on the pointwise evaluation of the convex conjugate of $u_\theta$ composed with $F$, or explicit, through an additional network $M_\xi$ trained to map samples $x$ from $\rho$ to $\nabla u_\theta^*\circ F(x)$.
    \item Because $M$ is not, in general, injective, we consider the ill-posed problem of inverting $M$: we approximate a stochastic map $I_\psi$, parameterized as a generator, that can generate inputs $x$ such as $M(x)=y$ for a given $y$. We use bridge matching for this task~\citep{debortoli2023augmented}.
    \item We use our approach to factorize gradients of surfaces in low dimensions and show how to use our tools to study the critical points of a non-convex energy $g$. Factorizing $G:=\nabla g$ as $\nabla u_\theta \circ M_\xi$, and estimating the stochastic map $I_\psi$ corresponding to $M_\xi$, our goal is to generate zeros of $\nabla g$. The minimizer of $u_{\theta}$ being $\nabla u_{\theta}^*(0)$ by definition of convex duality, the points generated as $I_\psi(\nabla u_{\theta}^*(0), \mathbf{z})$ where $\mathbf{z}$ is a Gaussian noise of suitable size should, in principle, result in points that are roots of $\nabla g$. We use the cross-entropy loss of a small MNIST digits LeNet~\citep{lecun1998gradient} classifier and show the ability to sample new parameters with low gradient and good performance on the recognition task.
\end{itemize}

\section{Background}\label{sec:back}
This section introduces neural methods that have been proposed to learn \citeauthor{monge1781memoire} maps between two distributions and recalls the polar factorization theorem in its original form.

\subsection{Neural Approaches to the Monge Problem}\label{subsec:monge}
The \citeauthor{monge1781memoire} formulation of the OT problem between two probability measures $\mu$ and $\nu \in \cali{P}(\R^d)$ seeks a map $T : \R^d \rightarrow \R^d$ that transports $\mu$ onto $\nu$, while minimizing the following transport cost:
\begin{equation} 
    \label{eq:monge}
    \cali{W}_2^2(\mu,\nu) := \inf_{\substack{T:\mathbb{R}^d\rightarrow\mathbb{R}^d\\T_{\#}\mu = \nu}} \int_{\R^d} \tfrac12 \|x-T(x)\|^2 \mathrm{d}\mu(x) 
\end{equation}
The existence of an optimal map $T^\star$ is guaranteed under fairly general conditions~\citep[\S1]{San15a}, when e.g. $\mu$ has a density w.r.t. the Lebesgue measure. In that case, \citeauthor{Brenier1991PolarFA}'s most famous theorem states that the Monge problem \eqref{eq:monge} has a unique solution, found at the gradient of a convex function $f^\star$ i.e. $T^\star = \nabla f^\star$. That convex function $f^\star$ is itself the solution of the following dual objective:
\begin{equation}\label{eq:dual} f^\star \in \underset{{f \in \mathrm{L}^1(\mu)}}{\arg \inf} \int_{\R^d} f \mathrm{d} \mu + \int_{\R^d} f^* \mathrm{d} \nu\,\,
\end{equation}
where the $f^*$ is the convex conjugate of $f$,
\begin{equation}\label{eq:convconj}
f^*(y) := \sup_{x\in\mathbb{R}^d} \langle x, y\rangle - f(x)\,.
\end{equation}
Note that the star symbol $*$ used for convex-conjugacy should not be confused with the star symbol $\star$, used throughout the paper to denote an optimal solution.
The OT map from $\nu$ to $\mu$ is also given by the inverse of $\nabla f^\star$ when it exists, $\nabla (f^\star)^*$.
The goal of neural OT solvers is to estimate $f^\star$ using samples drawn from the source $\mu$ and the target distribution $\nu$. \citet{pmlr-v119-makkuva20a,korotin2020wasserstein2} have proposed methods that build on input convex neural networks (ICNN), as originally proposed by~\citet{pmlr-v70-amos17b}, to parameterize the potential $f$ as an ICNN.
The main difficulty in these methods lies in handling the Legendre transform in~\eqref{eq:dual} of the ICNN variable.  
To address this difficulty, surrogate networks can be used to replace $f^*$, and we refer to~\citet{amos2023amortizing} for the most recent proposal to refine these implementations using amortized optimization.
Neural solvers have been used successfully in various applications, notably in single cell genomics \citep{Bunne2021.12.15.472775,NEURIPS2022_2d880acd}; see also~\citep{huang2020convex,cohen2021riemannian}. 

\subsection{Polar Factorization}
Given a probability distribution $\rho$ supported on a bounded set $\Omega$, \citeauthor{Brenier1991PolarFA}'s polar factorization theorem states that any vector field $F:\Omega  \rightarrow \R^d$ can be written as the composition of the gradient of a convex function $\nabla u:\Omega  \rightarrow \R^d$ with a map $M:\Omega  \rightarrow \Omega$ that preserves the distribution $\rho$ (ie $M_{\#}\rho = \rho$). In that decomposition, $\nabla u$ is the unique OT map from Brenier's theorem that transports the measure $\rho$ on $F_{\#} \rho$, since $F_{\#} \rho = (\nabla u \circ M)_{\#} \rho = \nabla u_{\#} (M_\#\rho)= \nabla u_{\#}\rho$. %
 
\begin{theorem}[\citeauthor{Brenier1991PolarFA} polar factorization]\label{theo:brenier}
Let $\rho$ be a probability measure whose support, $\Omega \incc R^d$, is a bounded set and $F:\Omega \rightarrow \R^d$ a square-integrable vector field being non degenerate i.e. $\int_{\R^d} \|F\|^2 \mathrm{d} \rho < \infty$ and $\rho(F^{-1}(A)) = 0$ on Lebesgue negligible subsets $A$ of $\,\R^d$. Then, there exists a convex function $u: \Omega \rightarrow \R$ and a map $M:\Omega \rightarrow \Omega$ that is measure preserving, i.e. $M_{\#}\rho = \rho$, such that:
\begin{equation}\label{eq:polar}F = \nabla u \circ M\,.
\end{equation}
Both $M$ and $\nabla u$ are unique.
\end{theorem}

\section{Neural Polar Factorization (NPF)}
We describe our method to compute the approximate polar factorization of a field $F$, using i.i.d samples $(x_1, \dots, x_n)\sim \rho$ and their evaluations $(F(x_i))_i$. We first estimate the convex potential $u$ in the decomposition $F=\nabla u\circ M$ using an ICNN $u_\theta$ as an OT \citet{Brenier1991PolarFA} potential using an improved ICNN architecture. Next, we show that the measure-preserving map $M$ can be defined implicitly for any $x$, by evaluating the convex conjugate of $u_\theta$ on $F(x)$: This requires a call to a convex optimization routine at each evaluation. Thanks to our ICNN's strong convexity, the transform~\eqref{eq:convconj} is well-posed. To underline the link of that approach to estimate $M$ using $u_\theta$, we use the notation $M_\theta(x)$. Alternatively, we also propose to learn an amortized model for $M$, by learning a network $M_\xi$ trained on a regression task using paired data samples $\{(x_i, M_\theta(x_i))\}$.

\subsection{(Low-Rank + Diagonal) Quadratic Layers in ICNNs} \label{subsec:icnn_arch}
ICNNs provide a neural network parameterization of convex functions. We propose a modification of the original architecture presented in~\citep{pmlr-v70-amos17b}.
Our approach is inspired by the Gaussian initialization outlined in~\citep{Bunne2021.12.15.472775} and the low-rank quadratic layers presented in \citep{korotin2020wasserstein2}. The original ICNN was designed to re-inject the input vector $x$, transformed by an affine map, at every layer, as can be seen in~\citep[Equation 2, $y\rightarrow x$]{pmlr-v70-amos17b}. \citeauthor[\S B.2]{korotin2020wasserstein2} proposed instead to modify $x$ with multiple low-rank quadratic positive definite (PSD) forms. The PSD constraint ensures convexity of each entry, while the low-rank choice ensures a reasonable number of parameters. We propose quadratic PSD forms that incorporate a \textit{positive} diagonal plus low-rank matrices~\citep{saunderson,liutkus2017diagonal}:
$$
Q_{A,\delta}(x) := \|\delta \circ x\|^2 + \|A x\|^2 = x^T\left(\textrm{diag}(\delta)+ A^TA\right)x\,.
$$
The network has $L+1$ layers for $L\geq 1$; we have highlighted in blue the new PSD 
 (diagonal + low-rank) terms:
\begin{equation}
\begin{aligned}
z_{0} &= \sigma_0 \left ({\color{blue}{[ Q_{A^i_0,\delta^i_0}(x)]_i}} + B_{0} x + c_{0} \right),\\
z_{\ell+1} &= \sigma_\ell \left ({\color{red}{W_{\ell}}} z_{\ell} + {\color{blue}{[Q_{A^i_\ell,\delta^i_\ell}(x)]_i}} + B_{\ell} x + c_{\ell} \right), \\ 
z_{L+1} &= \sigma_L \left ({\color{red}{w_{L}}}^T
z_{\ell} + {\color{blue}{Q_{A_L,\delta_L}(x)}} + b_{L}^T x + c_{L} \right) \in\mathbb{R}\,\\
\qquad u_{\theta}(x) &= z_{L+1}
\end{aligned}
\end{equation}
In all layers above, the index $i$ spans $1,\dots,q$, where $q$ is the size of the state vectors $z_\ell\in\mathbb{R}^q$. This augmented ICNN is parameterized with the following family of parameters,  
\begin{equation}
\begin{aligned}
\theta = \big( &{\color{red}{W_{1:L-1}}}\in(\mathbb{R}_+^{q\times q})^L, {\color{red}{w_L}}\in \mathbb{R}^q_+, \\
&\left({\color{blue}{\delta^i_{0:L-1}}} \in (\mathbb{R}^d_+)^L, {\color{blue}{A^i_{0:L-1}}}\in(\mathbb{R}^{r\times d})^L\right)_{i=1\dots q},\\
&{\color{blue}{\delta_{L}}} \in \mathbb{R}^d_+, {\color{blue}{A_{L}}}\in\mathbb{R}^{r\times d},\\
& B_{0:L-1}\in (\mathbb{R}^{q\times d})^L, b_{L}\in \mathbb{R}^d,\\
&c_{0:L-1} \in (\mathbb{R}^q)^L, c_L\in\mathbb{R}\big)\,.
\end{aligned}
\end{equation}
The activation functions $\sigma_{\ell}$ are convex, non-decreasing non-linear and all parameters in {\color{red}{red}} in addition to all diagonal vectors ${\color{blue}{\delta}}$ must be non-negative to ensure convexity.

\subsection{Estimating the Convex Potential $u$}
Starting from the existence result outlined in~\eqref{eq:polar}, we recover, by applying the push-foward map $F$ on $\rho$, that
$$
F_\sharp \rho = (\nabla u \circ M)_\sharp \rho = \nabla u_\sharp (M_\sharp \rho) = \nabla u_\sharp  \rho\,.
$$
Since $u$ is a convex function, it optimally transports $\rho$ on $\nabla u_{\#} \rho$ in the \citeauthor{monge1781memoire} sense.
Therefore, the defining feature of $u$ is that $\nabla u$ is the Monge map from $\rho$ to $F_\sharp \rho$. We use \citeauthor{amos2023amortizing}' solver \citeyearpar{amos2023amortizing} to estimate the potential $u$ that pushes $\rho$ onto $F_{\#}\rho$, from the empirical measures $\rho_n := \frac{1}{n} \sum_{i=1}^n \delta_{x_i}$ and $F_{\#}\rho_n = \frac{1}{n} \sum_{i=1}^n \delta_{F(x_i)}$.
Using this solver consists of parameterizing the convex function $u$ as an ICNN $u_{\theta}$ following \S\ref{subsec:icnn_arch} and parameterizing $\nabla u^*$ directly by an auxiliary vector-valued network $V_{\phi}$. The auxiliary network $V_{\phi}$ is learned by minimizing the objective:

\begin{equation*}
    \cali{L}_{\textrm{convex-dual}}(\phi)=\frac{1}{n} \sum_{i=1}^n \| V_{\phi}(F(x_i)) - \nabla u_{\theta}^*(F(x_i)) \|^2
\end{equation*}

One can show, using \citeauthor{doi:10.1137/0114053}'s envelope theorem \citeyearpar{doi:10.1137/0114053}, that $\nabla u_\theta^*(y)$ is the maximizer of the convex conjugate~\eqref{eq:convconj} problem for $u$ at $y$,
$$
\nabla u_\theta^*(y) = \arg\sup_x \langle x , y \rangle - u_\theta(x)
$$
Because $u_\theta$ is strictly convex, we compute the optimal solution solving $u_\theta^*(F(x))$ with a conjugate solver, e.g. gradient ascent, (L)BFGS~\citep{liu1989limited} or ADAM~\citep{kingma2014adam}. We call a conjugate solver $\textrm{CS}$ any algorithm that, for a given pair $(u, y)$, outputs an approximation of $\nabla u^*(y)$. This results in the loss:
\begin{equation}
    \label{eq:conv_dual_loss}
    \cali{L}_{\textrm{convex-dual}}(\phi)=\frac{1}{n} \sum_{i=1}^n \| V_{\phi}(F(x_i)) - \textrm{CS}(u_{\theta}, F(x_i))\|^2
\end{equation}
In practice, the latter is initialized with the predictions of $V_{\phi}$, which considerably reduces the number of iterations required for the solver to converge when $V_{\phi}$ starts making correct predictions. The parameters of the network $u_{\theta}$ are then updated alternatively, by taking steps along the gradients of the original dual objective of~\citet{pmlr-v119-makkuva20a}:
\begin{equation}
    \begin{aligned}
    \cali{L}_{\textrm{Monge}}(\theta) = \frac{1}{n} \sum_{i=1}^n & u_{\theta}(x_i) + \langle V_\phi(F(x_i)), F(x_i) \rangle\\
    & - u_{\theta}(V_{\phi}(F(x_i)))
    \end{aligned}
\end{equation}

\subsection{Estimating the Measure-Preserving Map $M$}
In the polar decomposition of $F$, $\nabla u$ is tasked with transporting $\rho$ on $F_{\#} \rho$, the measure-preserving map $M$ ensures then that $F = \nabla u \circ M$. To express $M$ as a function of $F$ and $u$, one simply has to apply the inverse of $\nabla u$ on both sides. When $u$ is strictly convex, we simply rely on the identity $\nabla u^*\circ \nabla u=\mathrm{Id}$ to obtain:
\begin{equation}
    \label{eq:meas_pres}
    M = \nabla u^* \circ F
\end{equation}

\paragraph{Evaluating $M$ using a Conjugate Solver.}
Given a conjugate solver $\textrm{CS}$ and the estimate $u_{\theta}$ for the ground truth potential $u$, we can inject them in~\eqref{eq:meas_pres} to get an estimation $\textrm{CS}(u_{\theta}, F(x))$ of $M(x)$ for a given $x$. Since this estimation depends on $\theta$, we define that approximation as $M_\theta(x)$,
\begin{equation}\label{eq:m_theta}
    M_\theta(x):=\textrm{CS}(u_{\theta}, F(x))\,,
\end{equation}
with a slight abuse of notation, since $\theta$ should not be understood as a parameter parameterizing $M$, but instead defining it implicitly through $u_\theta$ and $\textrm{CS}$. 

\paragraph{Neural Estimation for $M$.}
While $M_{\theta}$ does indeed provide an estimate of $M$, it may be convenient to parameterize the measure-preserving map of interest as a neural network $M_{\xi}$, defined with an independent set of parameters $\xi$. 
Borrowing a page from amortized optimization~\citep{amos2023tutorial}, 
$M_{\xi}$ can be used to initialize the conjugate solver used to estimate $M_{\theta}$ or even replace it when $M_{\xi}$ is sufficiently accurate. Furthermore, the parameterization of $M$ by $M_{\xi}$ is sometimes necessary when, e.g., $F$ is only given on a few samples, and one wishes to evaluate $M$ at any point. The neural map $M_{\xi}$ is then trained to minimize the following mean-squared error: 
\begin{equation}\label{eq:lossxi}\cali{L}_{\textrm{preserving}}(\xi) = \frac{1}{n} \sum_{i=1}^n \| M_{\xi}(x_i) - \textrm{CS}(u_{\theta}, F(x_i)) \|^2\,.\end{equation}
Note that while the loss in~\eqref{eq:lossxi} resembles~\eqref{eq:conv_dual_loss}, the network $V_\phi$ takes the transported point $F(x_i)$ as an input, whereas $M_\xi$ is only given $x_i$.

\paragraph{Evaluating The Measure Preservation of $M$.} In both cases, $M_\theta$, as evaluated with a conjugate solver, or its independently evaluated neural counterpart $M_{\xi}$ should be measure-preserving. Indeed, we will use (as in Figure~\ref{fig:topo_chamomix_npf}, bottom center plots) any departure from the identity
\begin{align*}
    M_{\#}\rho &= (\nabla u^* \circ F)_{\#}\rho = (\nabla u^*)_{\#} (F_{\#}\rho) = \rho\,,
\end{align*}
as a way to assess the quality of our factorization.

\subsection{Sampling according to the pre-image measure $M_{\theta}^{-1}$}
Measure-preserving maps $M$ are not invertible in general~\citep{ryff1970measure}, a well-known example in 1D being the doubling map defined as $M(x) = 2x \text{ mod } 1$ that preserves the Lebesgue measure rescaled to the interval $[0,1]$. This non-invertibility is of particular interest in the optimization and sampling applications we propose. For a given $y$, our goal will therefore be to generate inputs $x$ such that $M_{\theta}(x) = y$. To this end, we learn a generative process to sample according to the posterior density 

\begin{equation}
   \pi_{\theta}(x|y) = \frac{\mathbf{1}_{y=M_{\theta}(x)}\rho(x)}{\int_x \mathbf{1}_{y=M_{\theta}(x)}\rho(x) \mathrm{d}x}. 
   \label{eqn:posterior}
\end{equation}

We rely on the augmented bridge matching procedure presented in \citet{debortoli2023augmented} to learn the drift of the stochastic differential equation (SDE) formulated in~\eqref{eq:sde} so that, on input $X_0 = y$, the generated samples $X_1$ be distributed according to $\pi_{\theta}(x|y)$~\eqref{eqn:posterior}.
\begin{equation} 
   \text{d}X_t = ({X}_{\psi}(X_0,X_t) - X_t)/(1-t) \text{d}t + \sigma \text{d}B_t
   \label{eq:sde}
\end{equation}
\citeauthor{debortoli2023augmented}'s approach refines the bridge matching procedures that have been recently used to solve inverse problems \citep{somnath2023aligned, liu2023i2sb, chung2024direct}, by augmenting the learnable part of the drift ${X}_{\psi}$ with the initial point $X_0$ of the SDE. This slight adjustment allows to correctly recover the coupling measure $(M_{\theta}, \mathrm{Id})_{\#} \rho$ from the paired samples $\{(x_i, F(x_i))\}_{i=1}^n$ when ${X}_{\psi}$ is parameterized using a multilayer perceptron trained according to \Cref{alg:inverse}.

\begin{algorithm}[H]
\begin{algorithmic}[1]
\STATE $u_{\theta}$ $\longleftarrow$ Trained ICNN s.t. $\nabla u_{\theta}{}_{\#} \rho \approx F_{\#}\rho$
\STATE Initialize ${X}_{\psi}$ 
\WHILE{not converged}
    \STATE Draw a sample $(x_i, F(x_i))$ 
    \STATE Compute $y_i = \textrm{CS}(u_{\theta}, F(x_i))$
    \STATE Sample $t \sim \cali{U}([0,1])$
    \STATE Sample $z_i \sim \cali{N}(0, I_d)$
    \STATE $x_t := (1-t)y_i + tx_i + \sigma(t(1-t))^{1/2} z_i$
    \STATE $\cali{L}_{\psi} \leftarrow \frac{1}{n} \| {X}_{\psi}(y_i, x_t) - x_i\|^2$
    \STATE Update $X_{\psi}$ using $\nabla \cali{L}_{\psi}$
\ENDWHILE
\end{algorithmic}
\caption{Training of ${X}_{\psi}$}
\label{alg:inverse}
\end{algorithm} 

The optimized network ${X}_{\psi}$ is then plugged in~\eqref{eq:sde} that we solve with Heun's method as implemented in $\texttt{diffrax}$~\citep{kidger2021on} using $S$ discretization steps. Given a sample $y$ from $M_{\theta}{}_{\#} \rho$, solving the SDE~\eqref{eq:sde} using $X_0=y$ allows to generate an output $X_1$ distributed according to the posterior density~\eqref{eqn:posterior}. To alleviate the notations, we call $I_{\psi}$ the generative process such that $I_{\psi}(y, \mathbf{z})$ is the output $X_1$ returned by the differential equation solver associated to~\eqref{eq:sde} on the input $X_0=y$ when the injected gaussian noise $\mathbf{z}$ has been drawn from $\mathcal{N}(0,I_d)^{\otimes S}$
$$I_\psi(y,\mathbf{z}) = \mathrm{SDE}({X}_{\psi}, y, \mathbf{z}),\; y \in \mathbb{R}^d,\, \mathbf{z}\in\mathbb{R}^{d\times S}.$$
To generate several inputs $x$ from $\pi_{\theta}(x| y)$, one only needs to inject different noises $\mathbf{z} \sim \mathcal{N}(0,I_d)^{\otimes S}$ in $I_\psi(y,\cdot)$, i.e.
$$I_\psi(y,\cdot)_{\#} \mathcal{N}(0, I_d)^{\otimes S} \approx \pi_{\theta}(\cdot| y).$$

\section{NPF to Study Non-Convex Potentials $g$}
In this section, we focus on the polar factorisation of the gradient field $\nabla g$, where $g$ is a non-convex function of interest. We show how computing the NPF of $\nabla g$ together with the inverse map $I_{\psi}$ can be used to explore the space of critical points of $g$.
\subsection{On using the Inverse Map $I_{\psi}$ of $\nabla g$}

\paragraph{NPF on $G = \nabla g$. }
Let $g: \R^d \rightarrow \R$ be a function of interest supported on a bounded set $\Omega \subset \R^d$. Assuming that $\nabla g: \R^d \rightarrow \R^d$ meets the requirements of~\eqref{eq:polar}, \citeauthor{Brenier1991PolarFA}'s polar factorization states the existence of a convex function $u$ and a measure-preserving $M$ that preserves the rescaled Lebesgue measure on $\Omega$, $\cali{L}_{\Omega}$ such that: 
$$\nabla g = \nabla u \circ M.$$
For a given vector $v$, the points in $\Omega$ whose gradient with respect to $g$ is equal to $v$ are all transported by $M$ on the same point $\nabla u^*(v)$ i.e. 

$$M \left(\{x \in \Omega: \nabla g(x) = v \}\right) = \left \{\nabla u^*(v)\right\}.$$

In particular, the critical points of $g$ are all mapped by $M$ onto the minimizer of the function $u$, which is $\nabla u^*(0)$. 
\paragraph{On Extracting the Critical Points of $g$.}
When the NPF of $\nabla g$ is learned, resulting in $u_\theta, M_\xi$ and $I_\psi$, composing $I_{\psi}$ with $\nabla u_{\theta}*$ provides an inversion process for $\nabla g$. Generating an input point $x_v$ whose gradient is $v$ can in fact be done by first sampling $\mathbf{z} \sim \mathcal{N}(0,I_d)^{\otimes N}$ and successively applying $\nabla u_{\theta}*$ and $I_{\psi}$ to $v$:

$$x_v = I_{\psi}(\nabla u_{\theta}^*(v),\mathbf{z}),\; \text{where}\; \mathbf{z} \sim \mathcal{N}(0,I_d)^{\otimes S}.$$

As a special case, sampling the critical points of $g$ is done by taking $v = 0$ in the above procedure with different noises $\mathbf{z}$. Note, however, that this convexification requires estimating the polar factorization of $\nabla g$ as well as the inverse map $I_{\psi}$ over the entire space $\Omega$, which is computationally expensive. To optimize the $g$ function, we propose instead to combine this method with the Langevin Monte Carlo (LMC) algorithm to correctly estimate the polar factorization of $\nabla g$ around the minimums of $g$. 

\subsection{A LMC Method Assisted by NPF.}
\paragraph{LMC algorithm}
Given a smooth log-concave density
\begin{equation}
    \pi(x) = \frac{e^{-g(x)}}{\int_{x \in \R^d} e^{-g(x)} \mathrm{d}x}\,,
\label{eq:langevin_density}
\end{equation}
 with $g : \R^d \longrightarrow \R$, the Langevin Monte Carlo algorithm can sample from $\pi$
by starting from $x^{(0)}$ to iterate
\begin{align*}
   x^{(k+1)} &= x^{(k)} - \gamma \nabla g(x^{(k)}) + \sqrt{2 \gamma} z^{(k)}, \; \; z^{(k)} \sim \cali{N}(0,I_d).
\end{align*}
When $g$ is non-convex, the LMC algorithm lacks guarantees \citep{bj/1178291835, cheng2018convergence, DALALYAN20195278}. In particular, when $g$ has multiple local minima, the generated samples are highly correlated as the particles originating from the LMC algorithm often get stuck in some basins. For this reason, the LMC algorithm has been combined with methods enabling global jumps between modes \citep{10.1214/19-AOS1916, gabrie2022adaptive} to sample multi-modal distributions.

\paragraph{Sampling with Known Polar Factorization for $\nabla g$.} 
In this paragraph, we assume that the polar factorization $(\nabla u, M)$ and stochastic inverse map $M^{-1}$ of $\nabla g$ are known. To sample the modes of $\pi(x) \propto e^{-g(x)}$ when $g$ is non-convex, one can run the LMC algorithm on the convex function $u$ and sample back using an inverse generator $M^{-1}$:
\begin{align*}
    & y^{(k)} = M(x^{(k)}) \\
   &y^{(k+1)} = y^{(k)} - \gamma \nabla u(y^{(k)}) + \sqrt{2 \gamma} z^{(k)} \\
   &x^{(k+1)} = M^{-1}(y^{(k+1)}, z^{(k+\tfrac{1}{2})}).
\end{align*}
The LMC step on $u$ allows to move along a new descent direction or exploration direction to reach $y_{k+1}$ while $M^{-1}$ randomly generates a point $x^{(k+1)} \in \Omega$ whose gradient for $g$ is $\nabla u (y_{k+1})$. This way, the neighborhoods of $g$'s critical points are uniformly sampled, and a particle does not get stuck in one minimum as $M^{-1}$ permits global moves between all the basins. Because it is difficult to differentiate a minimum from a saddle point or a maximum when sampling critical points using the polar factorization of $\nabla u$, this procedure should be combined with Langevin steps on $g$ to escape non-minimum critical points. The following paragraph details the sampling algorithm and complements it by showing how the polar factorization of $\nabla u$ can be learned while sampling. 

\paragraph{Unknown Polar Factorization for $\nabla g$}  
When the polar factorization is unknown, we propose an algorithm that learns the polar factorization of $\nabla g$ as well as the inverse map $I_{\psi}$ using the generated particle trajectories. The algorithm alternates between $N $ Langevin steps on $g$ and $N$ Langevin steps on $u_{\theta}$, while $M_{\theta}$ and $I_{\psi}$ allow to transition between the two spaces. \Cref{alg:seq} details the steps of the procedure. The notation $\text{ LMC } (u_{\theta}, \gamma, y_i^{(k)}, N)$ means that $N$ LMC steps are performed on the function $u_{\theta}$ with a time step of $\gamma$ starting from the point $y_i^{(k)}$.

\begin{algorithm}[H]
\begin{algorithmic}[1]
\STATE Initialize $u_{\theta}$ and $I_{\psi}$
\STATE Initialize the particles $\{x_i^{(0)}\}_{1 \leq i \leq n}$
\STATE $k \leftarrow 0$
\WHILE{$k < k_{max}$}
    \IF{$k \mod N = 0$}
        \STATE $y_i^{(k)} =M_{\theta}(x_i^{(k)})$
        \STATE $y_i^{(k+1)}  = \text{ LMC } (u_{\theta}, \gamma, y_i^{(k)}, N)$
        \STATE  $x_i^{(k+1)}  = I_{\psi}(y_i^{(k+1)}, \mathbf{z})\; \text{with}\; \mathbf{z} \sim \mathcal{N}(0,I_d)^{\otimes S}$
    \ELSE
        \STATE $x_i^{(k+1)}  = x_i^{(k)} - \gamma \nabla g(x_i^{(k)}) + \sqrt{2 \gamma} z_i^{(k)}$
    \ENDIF
\STATE Update $u_{\theta}, I_{\psi}$ with $\{(x_i^{(k)}, \nabla g(x_i^{(k)}))\}_{1 \leq i \leq n}$
\STATE $k \leftarrow k+1$

\ENDWHILE
\end{algorithmic}
\caption{LMC-NPF}
\label{alg:seq}
\end{algorithm} 

The main insight of the proposed sampling algorithm is that LMC steps permit the exploration of the space locally, while NPF provides and stores a more global viewpoint, that is able to propose moves to potentially worthy areas.

\section{Experiments}
\subsection{Accuracy Metrics for NPF}

\paragraph{Assess NPF's Accuracy.}
\label{subsec:met}
When a field $G$ is only available through samples, the following three criteria, evaluated on unseen samples (or test set) $\{(x_j, G(x_j))\}_{1\leq j \leq m}$, are used to assess whether the estimated polar factorization is correct. 
\begin{itemize}[leftmargin=.3cm,itemsep=.0cm,topsep=0cm,parsep=2pt] 
\item To measure that the distributions $\nabla u_{\theta}{}_{\#} \rho$ and $G_{\#}\rho$ are close, we compute the Sinkhorn divergence $S_{\varepsilon}$~\citep{ramdas2017wasserstein,genevay2018learning,peyré2020computational} between the two point clouds $(G(x_j))_{1\leq j \leq m}$ and $(\nabla u_{\theta}(x_j))_{1\leq j \leq m}$. To quantify the scale of that measurement, we compare it with the distance between two batches of fixed size drawn from $(G(x_j))_{1\leq j \leq m}$. We also visualize this proximity by embedding the two point clouds using the TSNE algorithm~\citep{van2008visualizing} and superimpose them. 
\item The second criterion assesses whether $M_{\xi}$ is measure-preserving. Similarly, this is numerically estimated by computing the Sinkhorn divergence between the empirical measures associated with $(x_j)_{1\leq j \leq m}$ and $(M_{\xi}(x_j))_{1\leq j \leq m}$, and visualized with a TSNE embedding.
\item Finally, we evaluate the $L_2$ distance between $G$ and $\nabla u_{\theta} \circ M_{\xi}$ using the test set. Note that when $M_\theta$ is used (rather than $M_\xi$), that criterion is not useful since it only assesses the quality of the conjugate solver. 
\end{itemize}
\paragraph{Assess the Generative Inverse Map $I_\psi$.}
Given $y$, we should be able to sample among the antecedents of $y$ by $M_{\theta}$ using the multivalued map $I_{\psi}$. To quantify that, we estimate the average distance between the probability associated to the density $\pi_{\theta}(x|y)$~\eqref{eqn:posterior} and $I_\psi(y,\cdot)_{\#} \mathcal{N}(0, I_d)^{\otimes S}$ from samples. Given the finite test set $\{(x_j, M_{\theta}(x_j))\}_{1\leq j \leq m}$, it is unlikely to find a multitude of points with the same image. For this reason, we approximate $M_{\theta}^{-1}(M_{\theta}(x_k))$, by constructing the set
$$\mathcal{B}_{\alpha}(x_k)=\left\{x_j:\|M_{\theta}(x_j)-M_{\theta}(x_k)\|_2 \leq \alpha \right\}$$
and choose $\alpha$ such that the cardinal of $\mathcal{B}_{\alpha}(x_k)$ is $128$. We then compute the sinkhorn divergence between the predictions of $I_{\psi}$ on $M_{\theta}(\mathcal{B}_{\alpha_k}(x_k))$ and $\mathcal{B}_{\alpha_k}(x_k)$ and average it over all the $x_k$. We compare the obtained value with the distance of two batches of fixed size drawn independently from the $(x_j)_{1\leq j \leq m}$. We also evaluate the fact that the stochastic map $I_{\psi} \circ \nabla u_{\theta}^*$ approximates $G^{-1}$ by computing the quantity, using MC samples for $\mathbf{z}$,
\begin{equation}
    \frac{1}{m} \sum_{j=1}^m \E_{\mathbf{z} \sim \mathcal{N}(0,I_d)^{\otimes S}} \text{cosine}(G\circ I_{\psi}(\nabla u_{\theta}^*(G(x_j)), \mathbf{z}), G(x_j))
    \label{inv}
\end{equation}
which relies on the cosine similarity metric defined as $\text{cosine}(x,y) = \frac{\langle x, y \rangle}{\|x\| \|y\|}.$
\paragraph{Sinkhorn Divergence Parameter $\varepsilon$}
The $\varepsilon$ parameter used to compute sinkhorn divergence must be adapted to the scale of the data under consideration. In all our experiments, $\varepsilon$ is set to $\varepsilon = 0.05 \; \mathbb{E}_{x,x' \sim \rho} \|x-x'\|^2$ when computing divergence in the source space (ie to assess the accuracy of $M_{\xi}$ or $I_{\psi}$). Similarly, $\varepsilon$ is set to $\varepsilon = 0.05 \; \mathbb{E}_{x,x' \sim \rho} \|G(x)-G(x')\|^2$ when computing divergence in the target space (ie to assess the accuracy of $u_{\theta}$). In practice, the expectations are approximated using $2048 \times 2048$ MC samples.
\subsection{NPF of Topographical Data}
\paragraph{Dataset.}
We use the Python package $\texttt{elevation}$ to get the elevation of three regions of the world: Chamonix, London, and Cyprus. We estimate the gradients associated with the elevation in these regions with finite-differences, and obtain three datasets composed of (latitude, longitude) points paired with their gradients. We learn the polar factorization $(\nabla u_{\theta}, M_{\xi})$ of the underlying gradient field as well as the inverse map $I_{\psi}$. Because in these examples, $G$ is only given through samples, it can be interesting to parameterize the measure-preserving map using a neural network $M_{\xi}$. To assess the quality of our method NPF, we used a $85\%$ training / $15 \%$ test split. More details can be found in the appendix.

\begin{table}[h]
    \centering
    \begin{tabular}{ |p{5.4cm}||p{1.4cm}|p{1.4cm}|p{1.4cm}|  }
 \hline
  $ $ & Chamonix \\
 \hline
$S_{\varepsilon}$($\nabla u_{\theta}{}_{\#} \rho_{2048}$, $G_{\#} \rho_{2048}$) & 0.36 \\
 $S_{\varepsilon}$($ G_{\#} \rho_{2048}$, $G_{\#} \rho_{2048}'$) &   0.31 \\
 $S_{\varepsilon}(M_{\xi}{}_{\#} \rho_{2048},\rho_{2048}$)&   0.0022\\
 $S_{\varepsilon}$($\rho_{2048}, \rho_{2048}'$)&   0.0015 \\
 
 $\E_{x \sim \rho_n} \left[\|G(x) - \nabla u_{\theta} \circ M_{\xi}(x) \|_2\right]$ & 0.96\\
 $\E_{x, \mathbf{z}} \left[S_{\varepsilon}((I_{\psi}(M_{\theta}(\mathcal{B}_{\alpha}(x)), \mathbf{z}),\mathcal{B}_{\alpha}(x))\right ]$  &0.046 \\
 $S_{\varepsilon}$($\rho_{\ell}$, $\rho_{\ell}'$) & 0.039 \\
 \hline
\end{tabular}
    \caption{Polar factorization and Inverse multivalued map metrics for learning the gradient of the elevation in Chamonix area. For these metrics, $\rho_n$ and $\rho_n'$ are two empirical measures created from $n=2048$ samples drawn independently from the test set. Likewise, $\rho_\ell$ and $\rho_\ell'$ are two empirical measures created from $\ell=128$ samples.}
    \label{fig:tab_topo}
\end{table}

\begin{figure*}[h]
    \centering
    \begin{tabular}{cccc}
\includegraphics[scale=0.057, trim = 0cm 0cm 1cm 0cm, clip]{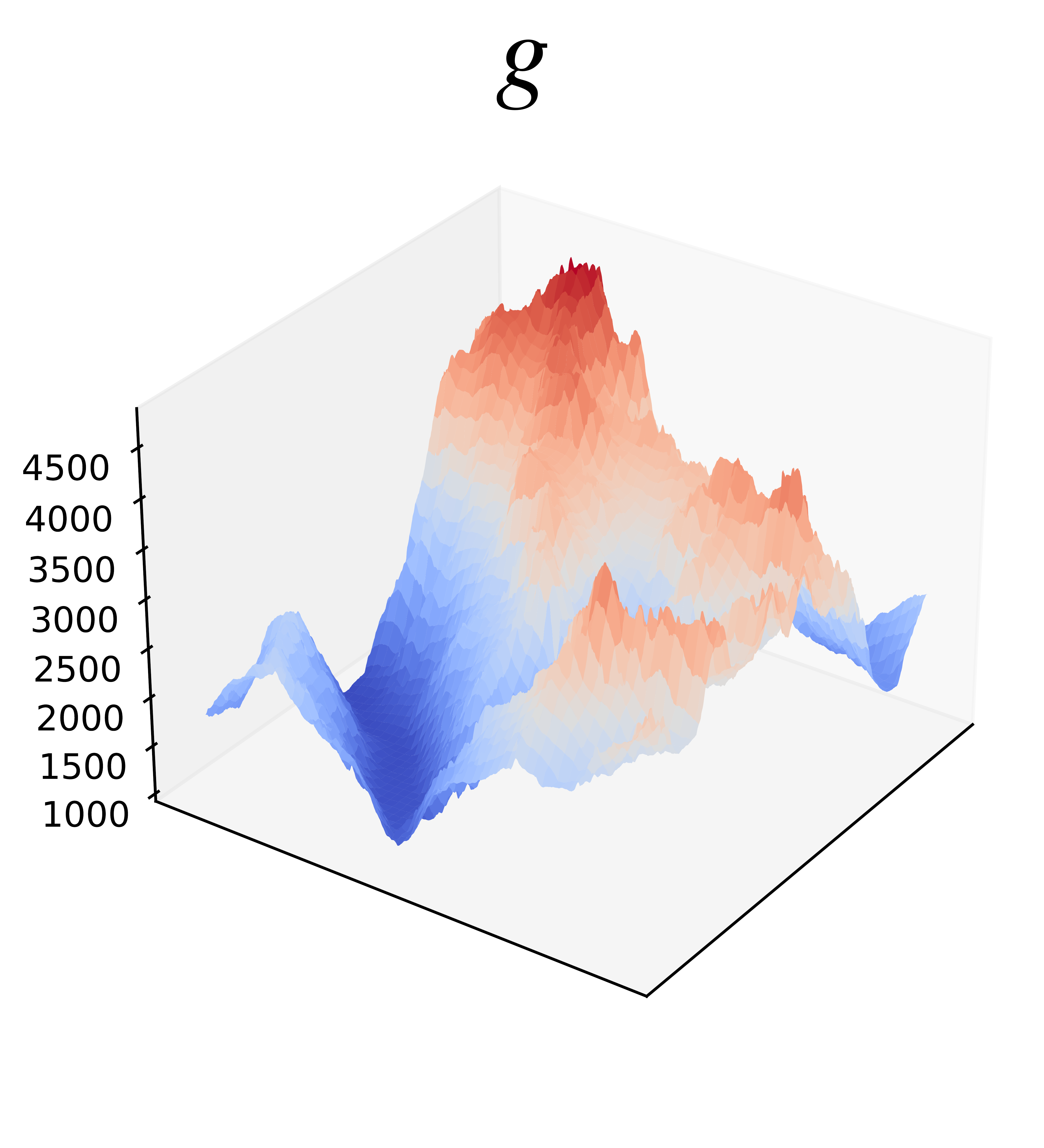}& \includegraphics[scale=0.057, trim = 0cm 0cm 1cm 0cm, clip]{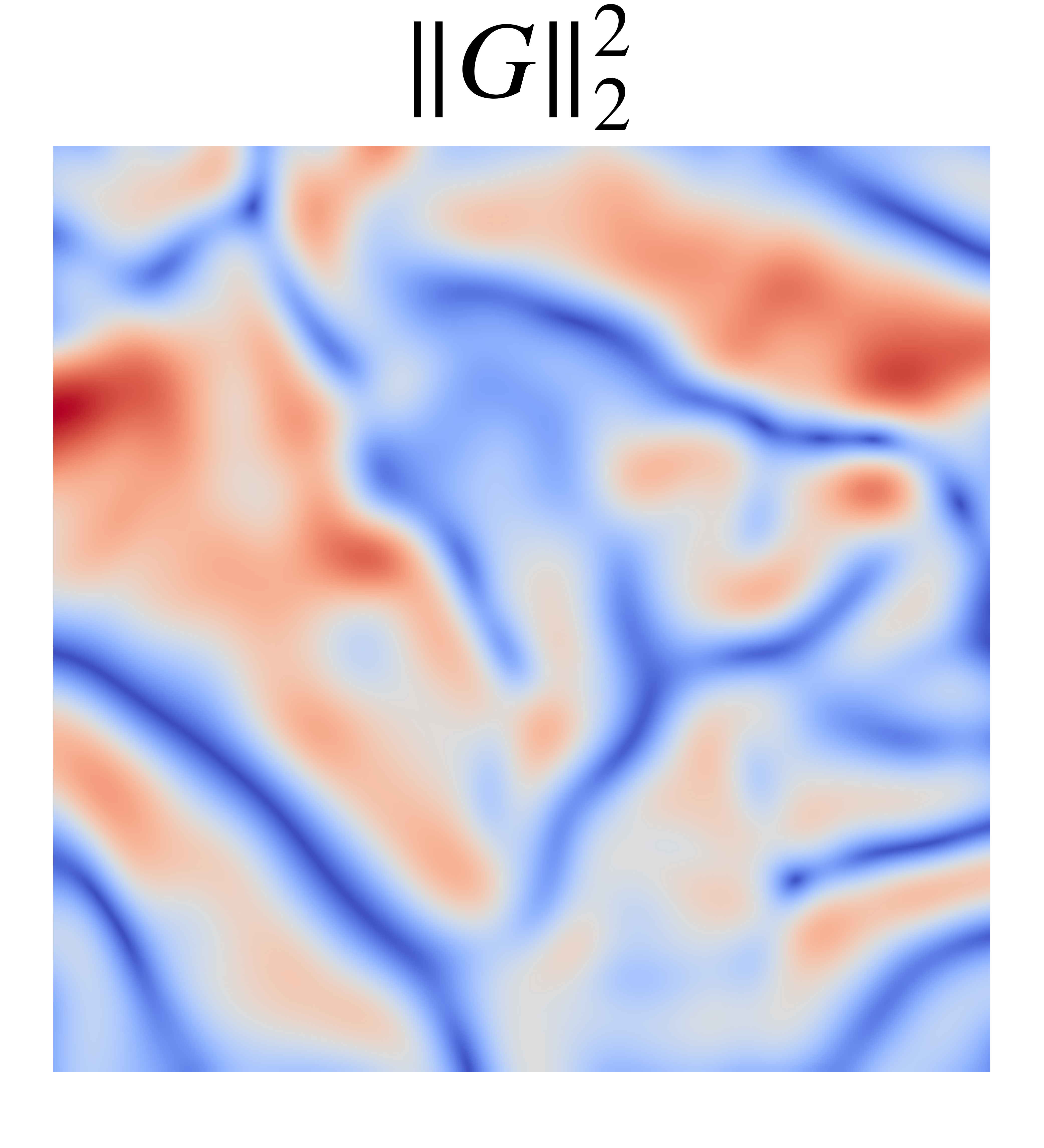} &
\includegraphics[scale=0.057, trim = 0cm 0cm 0cm 0cm, clip]{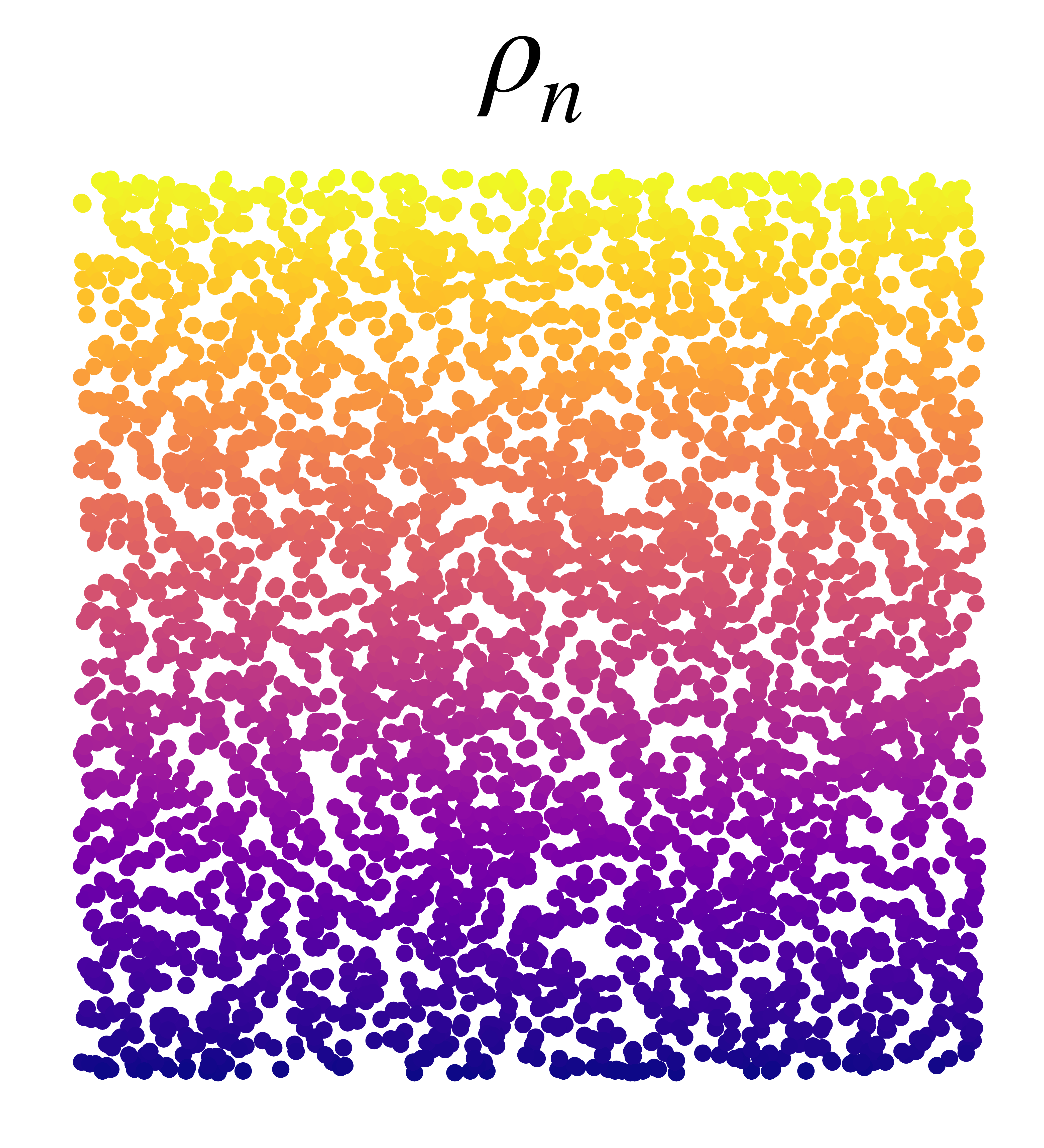} &
\includegraphics[scale=0.057, trim = 0cm 0cm 0cm 0cm, clip]{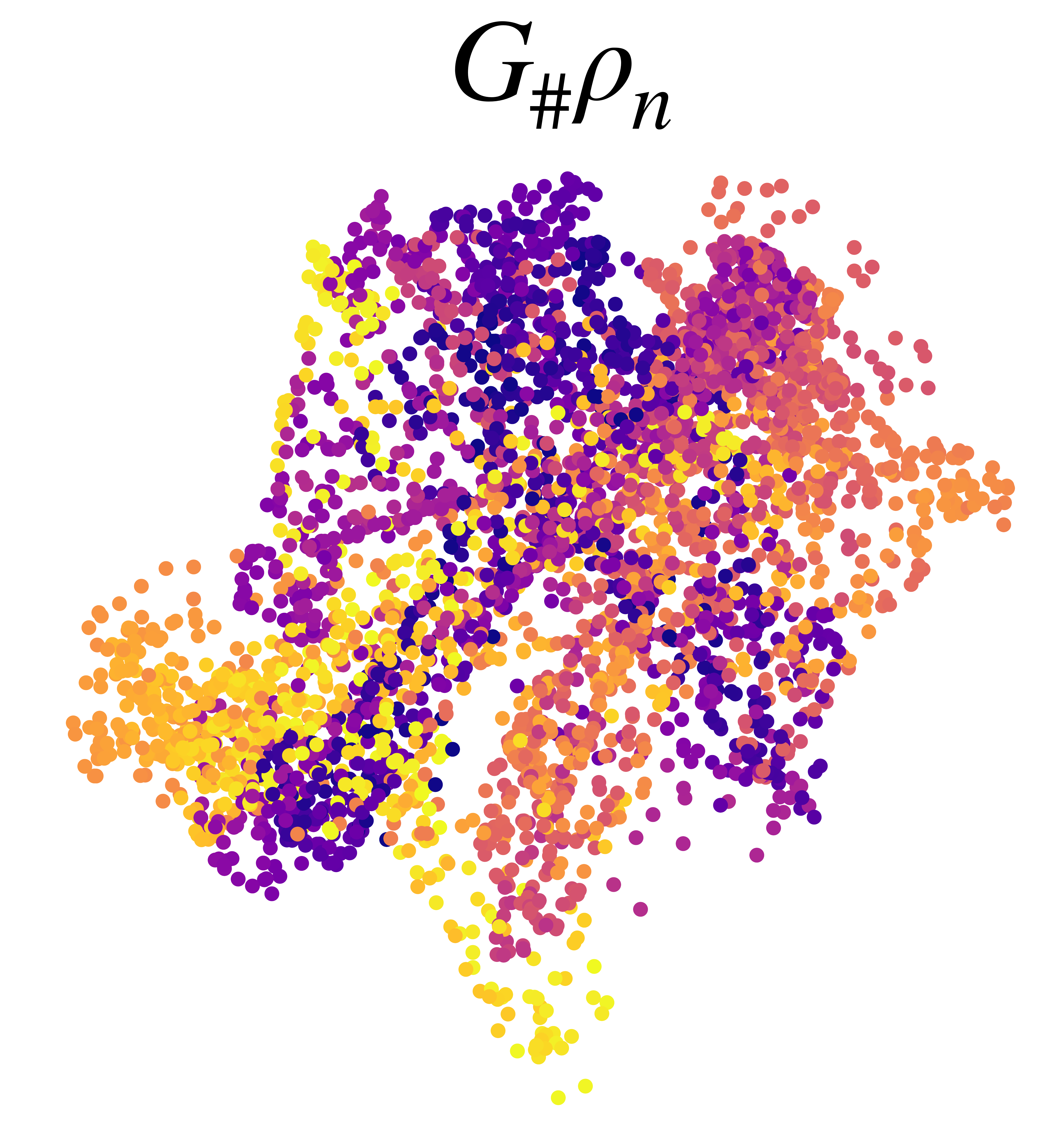} \\
\includegraphics[scale=0.057, trim = 0cm 0cm 0cm 0cm, clip]{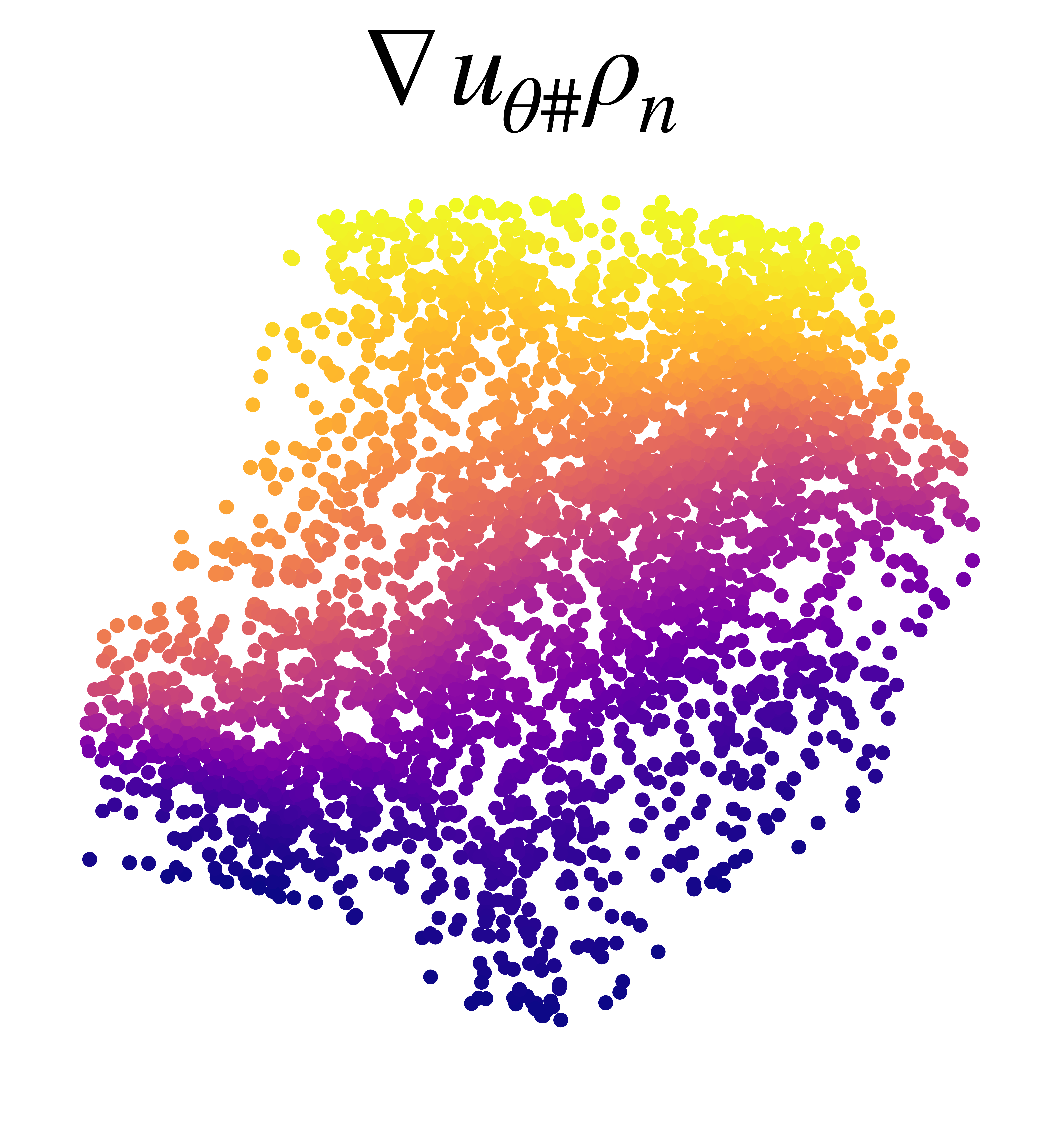} &
\includegraphics[scale=0.057, trim = 0cm 0cm 0cm 0cm, clip]{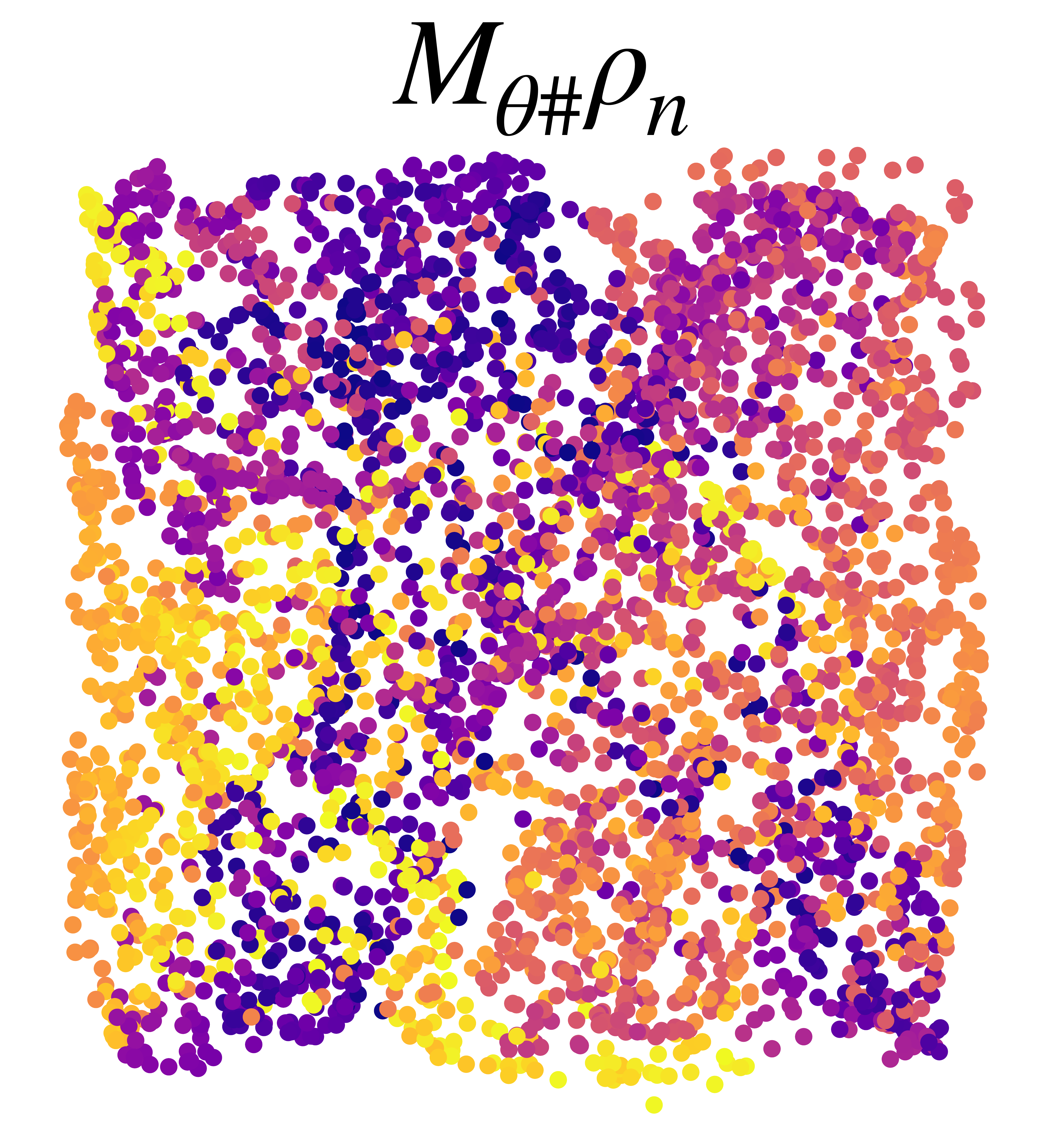} &
\includegraphics[scale=0.057, trim = 0cm 0cm 0cm 0cm, clip]{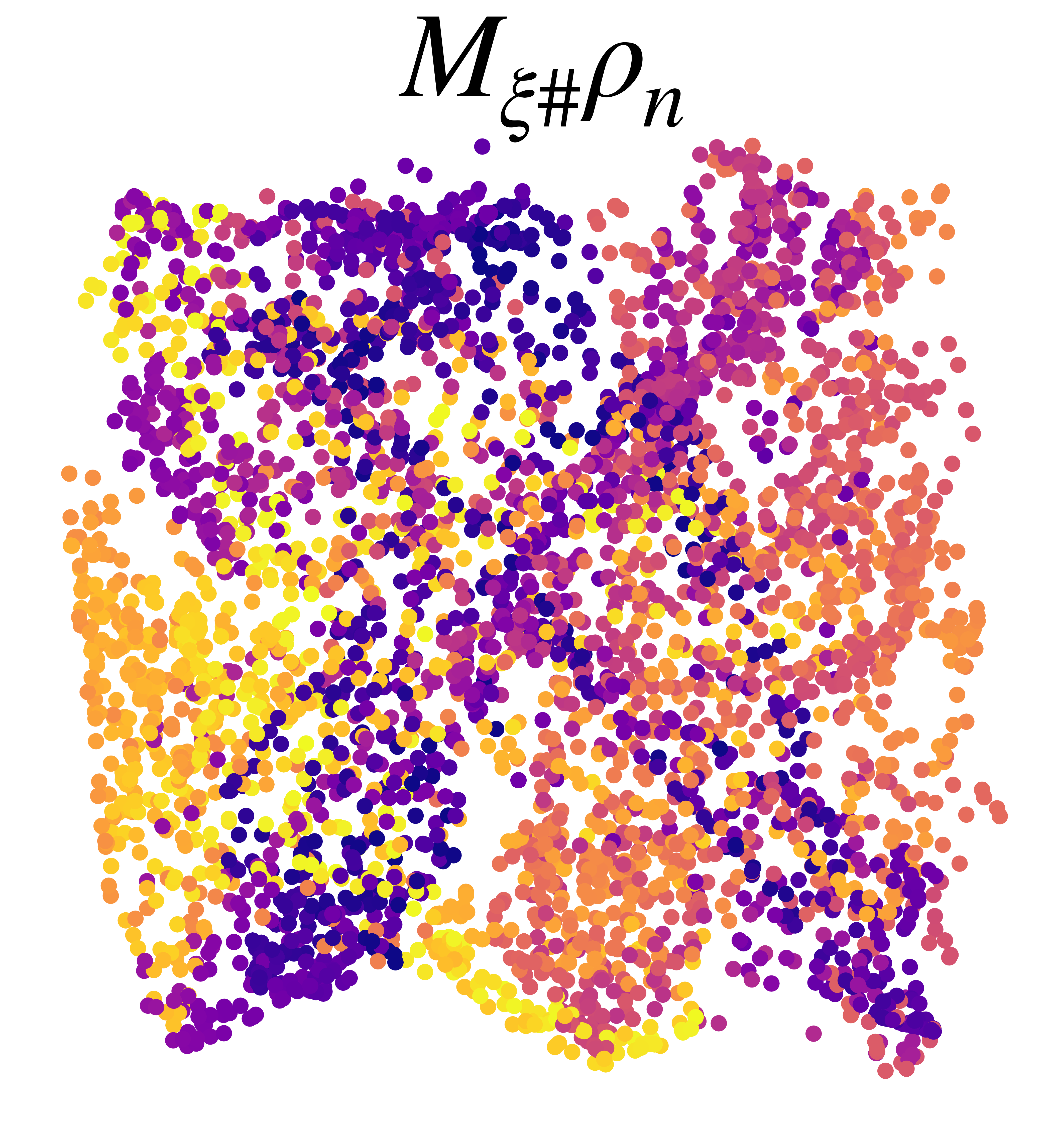} &
\includegraphics[scale=0.057, trim = 0cm 0cm 0cm 0cm, clip]{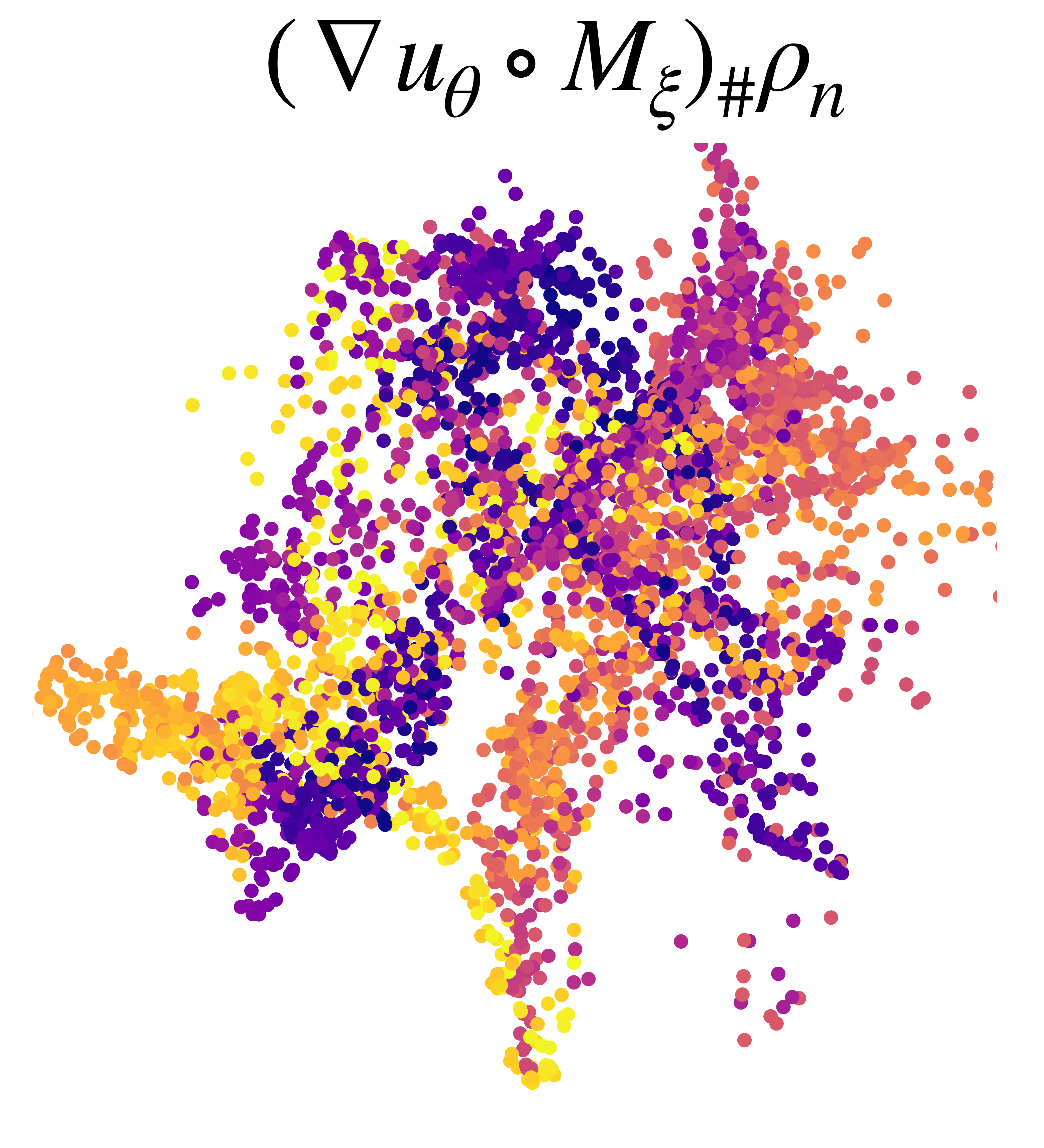}
\end{tabular}
    \caption{The $g$ function under study is the elevation in the Chamonix area (France). The figures show the respective action of the vector fields involved in the polar factorization of $\nabla g$ on a sample measure $\rho_n$. We observe that $\nabla u_{\theta} {}_{\#} \rho_n \approx G_{\#} \rho_n$. Both implicit and explicit measure-preserving maps $M_\theta$ \eqref{eq:m_theta} as well as the explicit network $M_{\xi}$ trained with the loss~\eqref{eq:lossxi} permutes the points of the distribution, ensuring that $G \approx \nabla u_{\theta} \circ M_{\xi}$ while $(M_\xi)_{\#} \rho_n \approx (M_\theta)_{\#} \rho_n \approx \rho_n$.}
    \label{fig:topo_chamomix_npf}
\end{figure*}

\paragraph{Polar Factorization Results.}
\Cref{fig:tab_topo} shows that the estimated NPF is accurate: the Sinkhorn divergence between the predicted distribution $\nabla u_{\theta}{}_{\#} \rho_n$ and the target distribution $G_{\#} \rho_n$ is of the same magnitude as the divergence between two batches $\rho_n, \rho_n'$ of the same size taken from the target. Similarly, the Sinkhorn divergence between $\rho_n$ and its image by $M_{\xi}$ is of the same order as the distance between two batches of same size drawn from the source. The reconstruction of $G$ is also quite satisfactory as corroborated visually (\Cref{fig:topo_chamomix_npf}).

\begin{figure}[h]
    \centering
    \begin{tabular}{c}

\includegraphics[scale=0.12, trim = 0cm 25cm 0cm 25cm, clip]{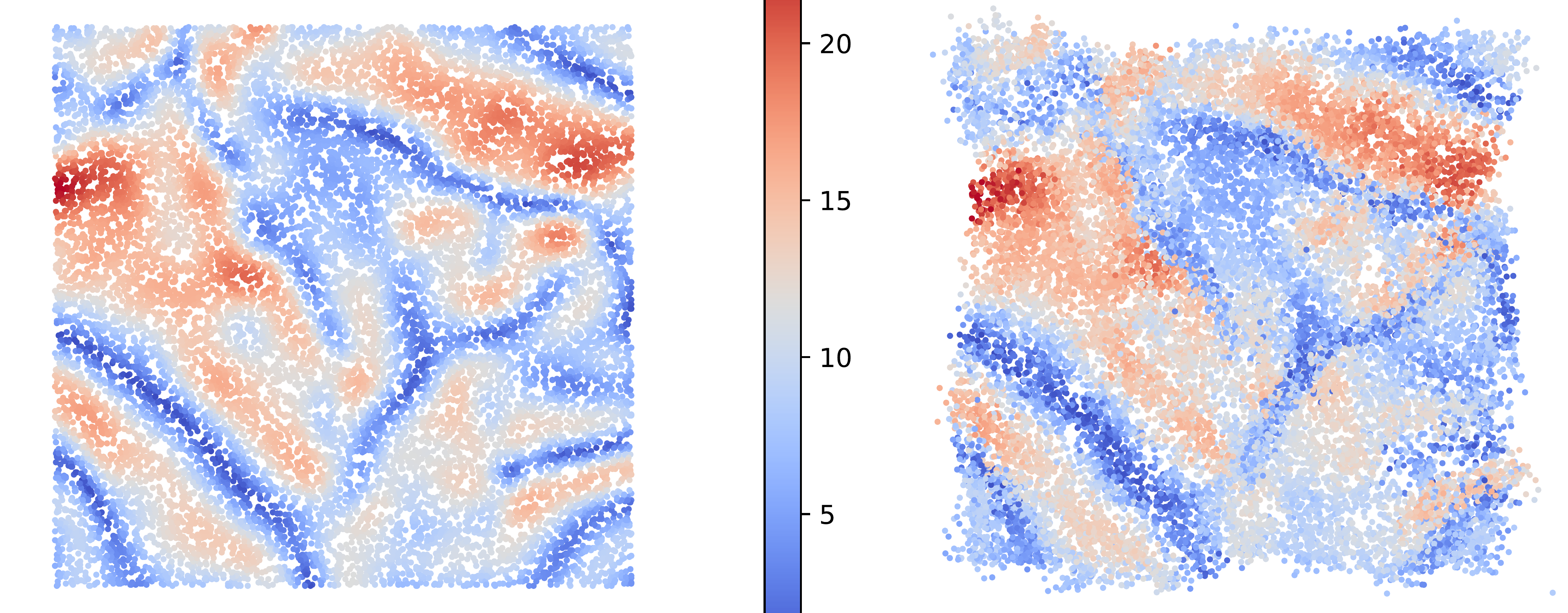}
\end{tabular}
    \caption{$I_{\psi}$'s ability to replace gradients in the original $\Omega$ space for the example of Chamonix region's elevation gradient. The figure on the right is generated by returning the gradients $\nabla g_{\#}\rho_n$ to their initial position in the image via $I_{\psi} \circ \nabla u_{\theta}^*$. This position is then colored according to the initial gradient norm (before transport). We can compare the result with the image on the left, generated by sampling uniformly in $\Omega$ space and colored according to the norm of their gradient.}
    \label{fig:topo_chamomix_inv}
\end{figure}

\paragraph{Inverse Map Results.}
The data from \Cref{fig:tab_topo} indicates that $I_{\psi}$ generates the antecedents of the images by $M_{\theta}$ accurately: the estimated quantity $\E_{x, \mathbf{z}} \left[S_{\varepsilon}((I_{\psi}(M_{\theta}(\mathcal{B}_{\alpha}(x)), \mathbf{z}),\mathcal{B}_{\alpha}(x))\right ]$  which is approximated using MC samples, is comparable to the distance between two batches of size $128$ drawn from the source distribution. To visualize these performances, we transported the samples $(G(x_j))_{1 \leq j \leq m}$, that store gradients of the elevation, using $I_{\psi} \circ \nabla u_{\theta}^*$ that should estimate the inverse generative map $G^{-1}$. We expect very high gradients to be sent to points where the elevation varies rapidly, such as the sides of mountains in the Chamonix example. To visualize where a gradient was sent, we plot a point at this localization and color it according to the norm of the gradient from which it originates. We compare the image generated by this process with the one obtained by coloring directly the points $(x_j)_{1 \leq j \leq m}$ using their associated gradients. In the three cases (Chamonix, London, Cyprus), the two images look quite similar (\Cref{fig:topo_chamomix_inv}), showing the quality of our reconstruction.

\subsection{Learn an NN Optimization Landscape using NPF}\label{subsec:npf}
In this experiment, we consider a minimal neural architecture capable of classifying MNIST digits. Inspired by the LeNet architecture~\citep{lecun1998gradient}, we use two convolutional layers, each followed by a Relu and a max pooling operation. A classification layer leads to an output layer of 10 neurons, followed by a softmax. The loss function is the cross entropy, computed with MNIST train dataset minibatches of size 128, and the vector field under study is the gradient of that loss for the $d=222$ parameters of the neural network. The loss landscape of a non-linear neural network being very chaotic \citep{NEURIPS2018_a41b3bb3}, we do not expect to learn the polar factorization of the associated gradient field perfectly over the all optimization space $\Omega$. The optimization space we are considering is $\Omega = [-1, 1]^{222}$.

\begin{figure}[h]
    \centering
    \begin{tabular}{cc}
\includegraphics[scale=0.055, trim = 0cm 0cm 0cm 0cm, clip]{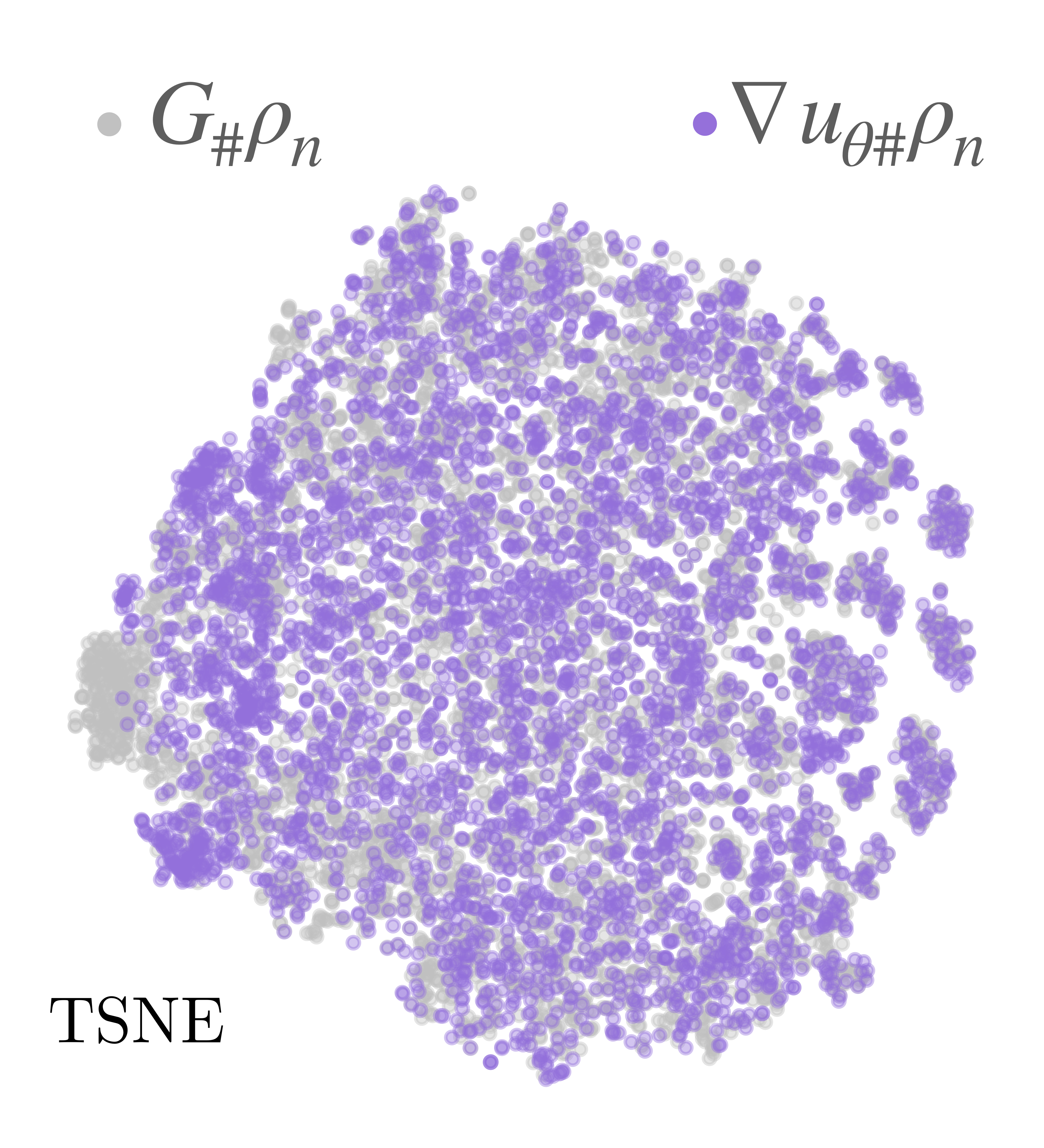} & 
\includegraphics[scale=0.24, trim = 0cm 0cm 0cm 0cm, clip]{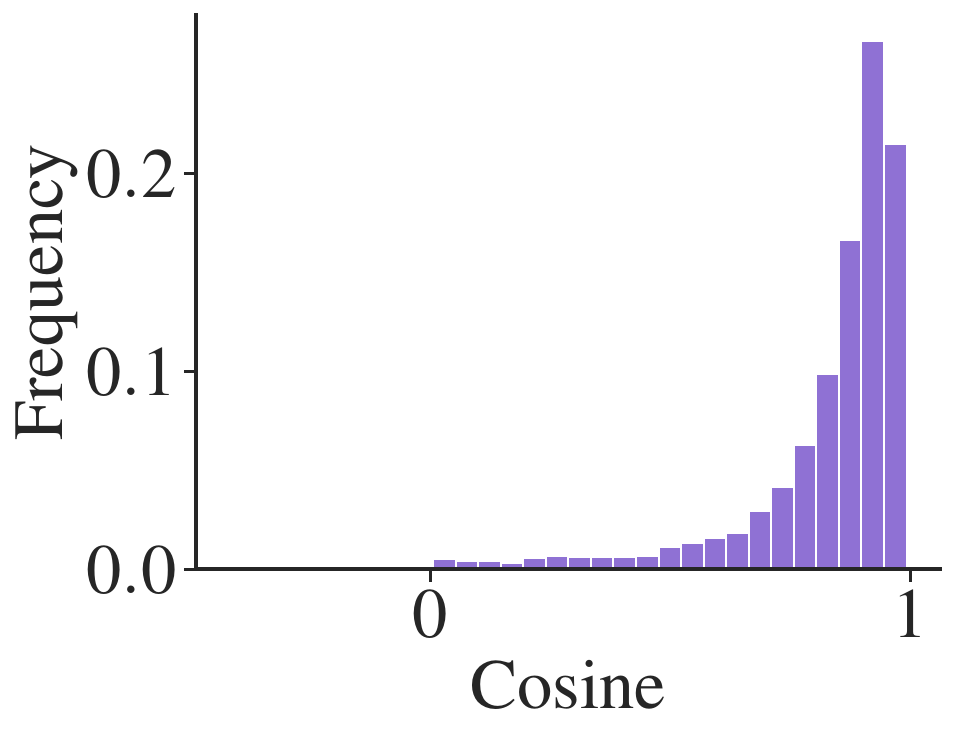}
\end{tabular}
    \caption{Performance of the NPF and the inverse map $I_{\psi}$ on the \S\ref{subsec:npf} MNIST classifier experiment where the loss gradient is learned over the entire space $\Omega$. The TSNE allows to visualize in 2D the overlap of the predicted distribution $\nabla u_{\theta}{}_{\#} \rho_n$ with the target distribution $G{}_{\#} \rho_n$ while the cosine similarity, mentioned in \S\ref{subsec:met}, shows that $I_{\psi}$ permits to accurately generate weights associated with a given gradient.}
    \label{fig:graph_perf_lenet_all}
\end{figure}

\paragraph{Polar Factorization and Inverse Map Results.}
According to \Cref{fig:tab_class}, we see that, overall, NPF manages to learn that vector field, but it lacks, as expected, accuracy in some parts of the space. This is, e.g., revealed visually using the TSNE plot from \Cref{fig:graph_perf_lenet_all}. Similarly, $I_{\psi }$ can be used to invert $M_{\theta}$ according to \Cref{fig:tab_class} and the histogram associated with the cosine similarity on \Cref{fig:graph_perf_lenet_all} confirms that we can choose a certain gradient $v$ and use $I_{\psi} \circ \nabla u_{\theta}^*$ to generate classifier weights whose gradient is approximately $v$. In particular, $I_{\psi}$ allows the generation of correct critical points (\Cref{fig:graph_all}), which, however, have a low accuracy. 
This is due to our uniform sampling procedure in the space $[-1, 1]^{222}$ that only reveals bad critical points with an accuracy of $10 \%$ which is the
performance of a classifier with random weights as shown in \Cref{fig:graph_all}.

\begin{table}[h]
    \centering
    \begin{tabular}{ |p{5.9cm}||p{1.6cm}| }
    \hline
$S_{\varepsilon}$($\nabla u_{\theta}{}_{\#} \rho_n$, $G_{\#} \rho_n$)  &  $97.4 \color{gray}\pm 6.5$\\
$S_{\varepsilon}$($G_{\#} \rho_n$, $G_{\#} \rho_n'$) & $93.6 \color{gray}\pm 6.2$\\
$\E_{k, \mathbf{z}} \left[S_{\varepsilon}((I_{\psi}(M_{\theta}(\mathcal{B}_{\alpha}(x_k)), \mathbf{z}),\mathcal{B}_{\alpha}(x_k))\right ]$ &  $107.3 \color{gray}\pm 0.7$  \\
 $S_{\varepsilon}$( $\rho_{\ell}$, $\rho_{\ell}'$)  &  $105.7 \color{gray}\pm 0.5$  \\
$\E_{y \sim G_{\#} \rho_n, \mathbf{z}} \text{cosine}(G\circ I_{\psi} (\nabla u_{\theta}^*(y), \mathbf{z}), y)$ & $0.81 \color{gray}\pm 0.20$ \\

 \hline
\end{tabular}
    \caption{Polar factorization and Inverse multivalued map metrics for learning the gradient of the MNIST classifier loss function. The metrics are computed using emprical distributions of respectively $n=2048$ and $\ell = 128$ samples.}
    \label{fig:tab_class}
\end{table}

\subsection{Learning NPF using Gradient Flow of Particles}\label{subsec:exp-mnist}
In \S\ref{subsec:npf}, we requested that NPF learns the entire gradient field. This of course limits the ability, given a certain budget of samples, to provide a good approximation of critical points through the inverse generative map. In particular, our uniform sampling procedure does not allow to reveal interesting critical points. In this experiment, we train NPF using gradient descent trajectories to focus on those areas. To do this, we initialize $1024$ particles randomly and have them follow a gradient flow. We use these trajectories' samples to learn NPF and sample critical points of the classifier.  

\begin{figure}[b]
    \centering
    \begin{tabular}{cc}
\includegraphics[scale=0.24, trim = 0cm 0cm 0cm 0cm, clip]{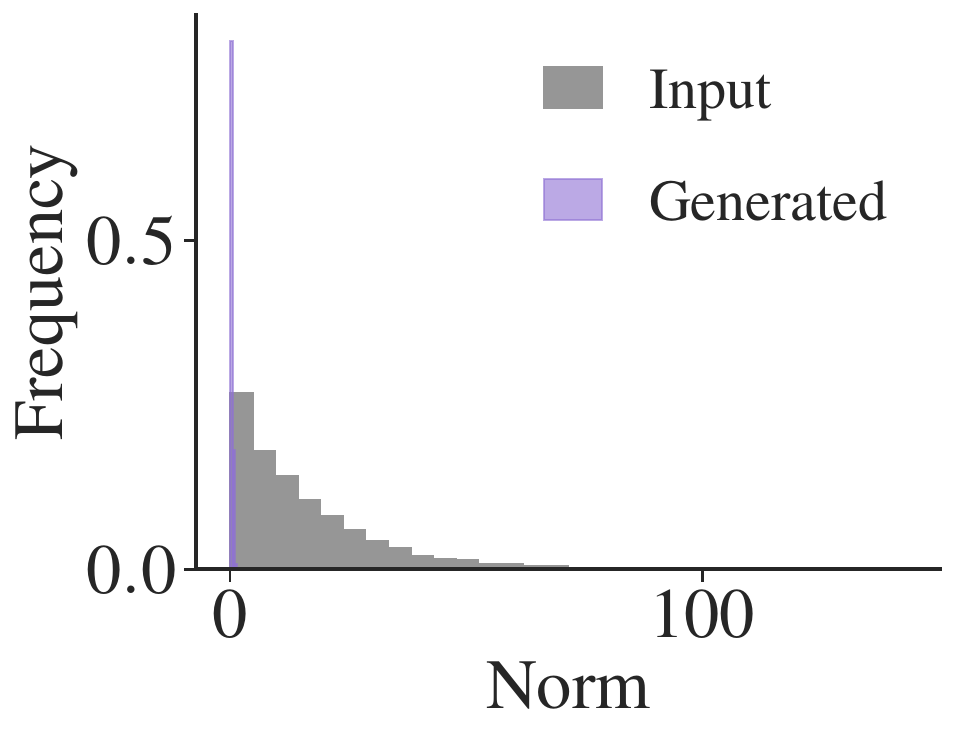} &
\includegraphics[scale=0.24, trim = 0cm 0cm 0cm 0cm, clip]{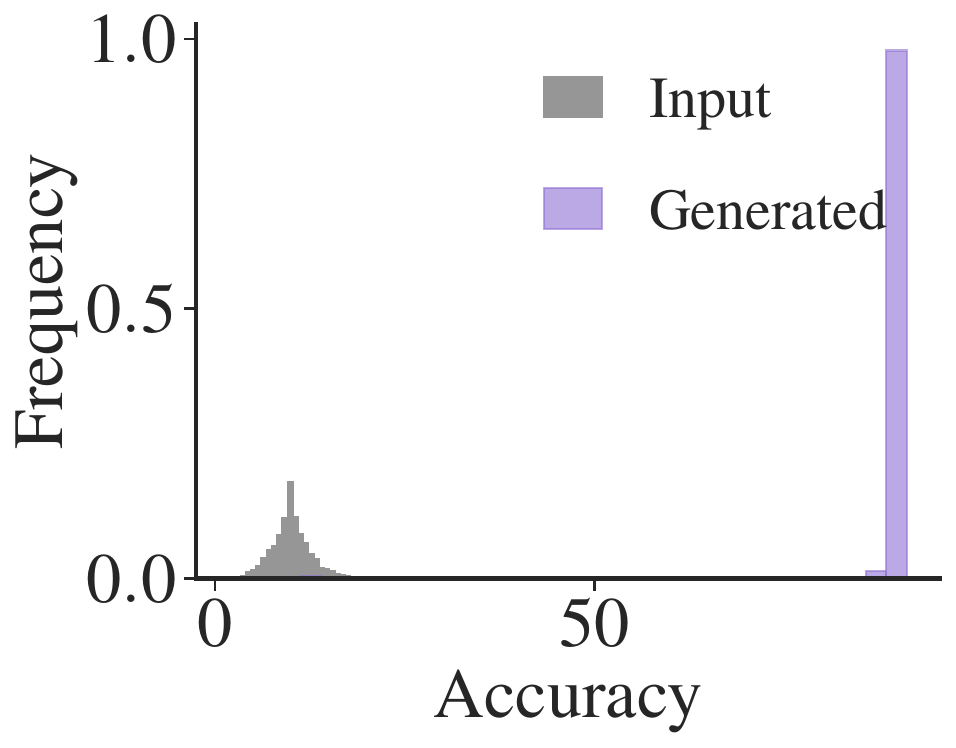}
\end{tabular}
    \caption{Characteristics of the points sampled using LMC-NPF for the \S\ref{subsec:lmc_npf} MNIST classifier experiment. In gray the classifier weights have been drawn uniformly in the $\Omega$ optimization space, while in purple the weights have been sampled using \Cref{alg:seq}. Generated samples are critical points which are good minima, as shown by the accuracy statistics.}
    \label{fig:graph_samp_crit_}
\end{figure}
\paragraph{Polar Factorization and Inverse Map Results.}
We observe that NPF is faithful near good accuracy basins: the Sinkhorn divergence between the distribution generated by $\nabla u_{\theta}$ and the target is of the same order as that between two batches of size $2048$. As for $I_{\psi}$, the gradients of the generated weights do have a norm close to $0$, and the cosine similarity distribution reveals that the direction of the gradients is globally learned (\Cref{fig:graph_desc_part}). Moreover, we can see in \Cref{fig:graph_ccp} that the critical weights generated contain a large part of valid minima. 

\subsection{LMC-NPF on MNIST}
\label{subsec:lmc_npf}
We use LMC-NPF (\Cref{alg:seq}) to sample the loss of the MNIST classifier considered in~\S\ref{subsec:npf}. Our sampling algorithm is preceded by a warm-up containing particle descents to explore good minima before sampling. 

\begin{figure}[h]
    \centering
    \begin{tabular}{c}
\includegraphics[scale=0.24, trim = 0cm 0cm 0cm 0cm, clip]{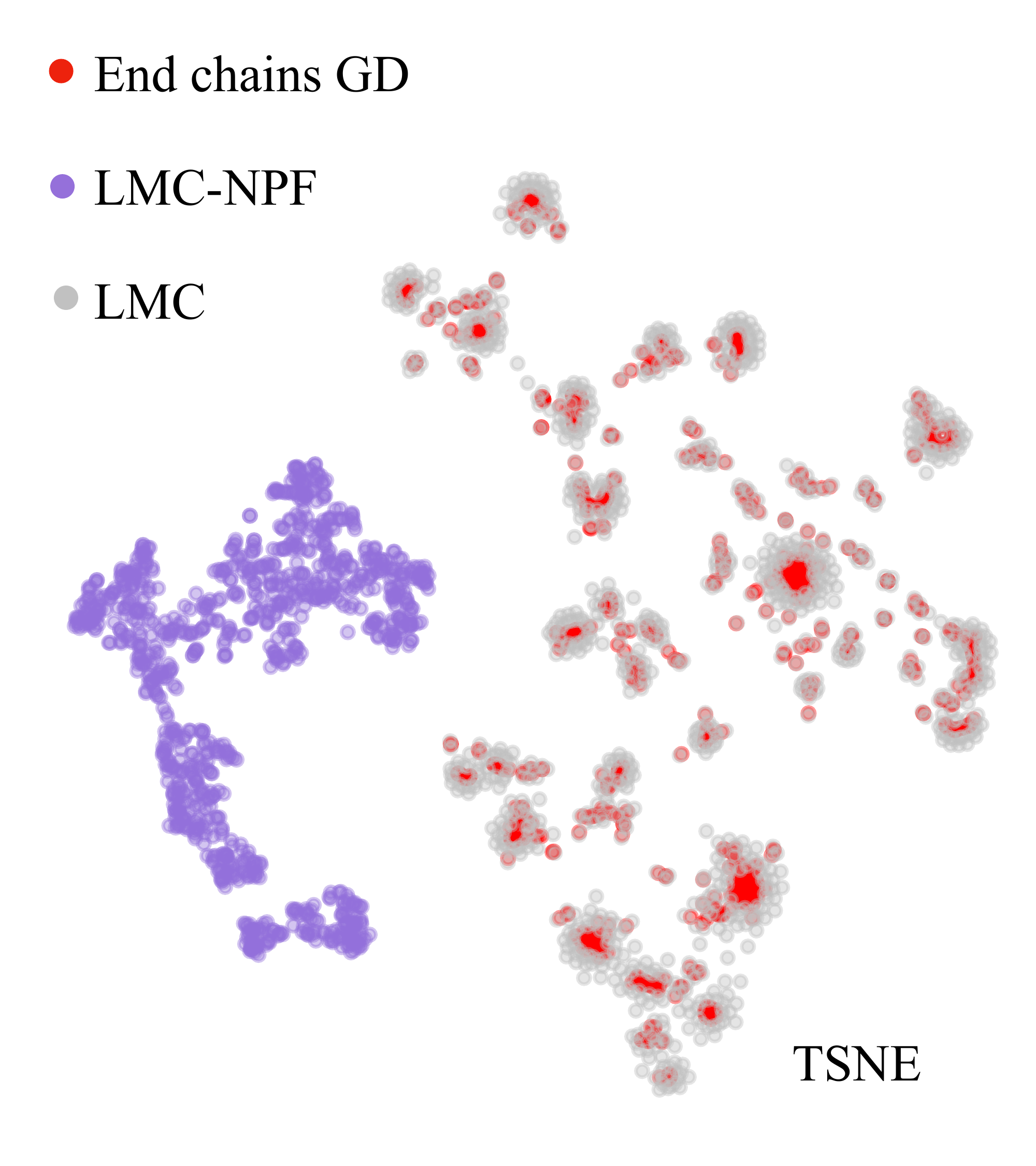}
\end{tabular}
    \caption{Results of LMC-NPF applied to the MNIST classifier loss function. The TSNE is used to represent the final particles resulting from descent trajectories during the Warm-up period (in red), the particles sampled by the LMC algorithm (in grey), and the particles sampled by LMC-NPF (in purple). }
    \label{fig:graph_samp_tsne}
\end{figure}

\paragraph{Sampling Algorithm Results.}
Following the warm-up, the TSNE (\Cref{fig:graph_samp_tsne}) as well as \Cref{fig:graph_samp_crit_} show that our sampling algorithm proposes high-accuracy weights that are completely different from the minima found during the warm-up period. Since LMC-NPF alternates between Langevin steps on $g$ and Langevin steps on $u_{\theta}$, we demonstrate that these minima had indeed been discovered through the use of NPF by running a LMC algorithm initialized with the final warm-up particles. The latter was parameterized the same way as the one used in LMC-NPF, with the same number of iterations. We observe that the LMC algorithm samples around the warm-up particles but does not detach itself from them. This confirms that the use of PFNet in the sampling procedure permits the discovery of new local minima.

\subsection{ICNNs Benchmark}
We compare three different ICNN architectures: the one proposed in \citet{amos2023amortizing}, the same
architecture but with an extra full quadratic layer at the end and ours (with a rank of $1$ for the intermediate quadratic layers) making sure that the architectures have a comparable number of parameters. In the first experiment, the starting measure is a $d$-dimensional standard Gaussian while the target is a gaussian of mean $0$ and covariance matrix $\text{diag}(1, 2, ..., d)$. The $\mathcal{L}_2$ unexplained variance percentage \citep{korotin2021neural} from \Cref{tab_icnn_bench_gauss} indicates that the architecture we propose provides a better estimation of the OT map for all the dimensions we considered. In the second experiment, the starting measure is still a $d$-dimensional standard gaussian and the objective is to map a mixture of gaussians whose modes have various sizes. The estimated Sinkhorn divergences between the generated distribution and the target (\Cref{tab_icnn_bench_multi}) shows that our proposed architecture is the one that best fits the multimodal distribution.

\section{Conclusion}
\citeauthor{Brenier1991PolarFA}'s polar factorization is arguably one of the most far-reaching results discovered in analysis in the last century, underpinning the better known \citeauthor{Brenier1991PolarFA} theorem on the existence of solutions to the~\citet{monge1781memoire} problem. We proposed in this work the first implementation, to the best of our knowledge, of that factorization that is applicable to higher-dimensional settings. To do so, we have used the recently proposed machinery of neural optimal transport solvers. Beyond simply exploiting this result, we have also proposed to estimate a multivalued map that approximates the inverse of the measure-preserving map component in the polar factorization. We have shown that such an inverse map can be of potential use to sample the optimization landscape of non-convex potentials.  An interesting direction for perfecting the sampling algorithm would be to reweight the samples according to their probability, in the same vein as SMC samplers~\citep{del2006sequential}. This would require knowledge of the probability distribution generated by the generative model $I_{\psi}$, which is not possible with the current methodology.

\section*{Acknowledgements}
This work was performed using HPC resources from GENCI–IDRIS (Grant 2023-103245). This work was partially supported by Hi! Paris through the PhD funding of Nina Vesseron.
\bibliography{References}
\bibliographystyle{plainnat}

\clearpage
\newpage
\appendix
\onecolumn
\section{Computation of the convex conjugate} 

Given a function convex function $u$, and a point $y$, the objective $J_u(x,y) = \langle y, x \rangle - u(x)$ is concave with respect to $x$ and $u^*(y) = \sup_{x} J_u(x,y)$ can be computed using optimization algorithms like gradient ascent, (L)BFGS or Adam. As for $\nabla u^*(y)$, taking the gradient for $y$ necessitates differentiate through a supremum. In our case, $u$ is strictly convex a.e. and the supremum becomes a maximum : 
$$u^*(y) = \max_{x} J_u(x,y)$$
\citeauthor{doi:10.1137/0114053}'s envelope theorem \citeyearpar{doi:10.1137/0114053} allows to differentiate through this maximum and to write:
\begin{align*}
    \nabla_y u^*(y) &= \nabla_y \max_{x} J_u(x,y) \\
    &= (\nabla_y J_u(x,y))(x^*(y))
\end{align*}
where $x^*(y)$ is the optimal $x$ that maximizes $ J_u(x,y)$. Because $\nabla_y J_u(x,y) = x$, we get that 
$$\nabla_y u^*(y) = x^*(y)$$ 

\section{Preconditionned LMC}
\begin{align*}
   x^{(k+1)} &= x^{(k)} - \gamma \nabla f(x^{(k)}) + \sqrt{2 \gamma} z^{(k)}, \; \; \; z^{(k)} \sim \cali{N}(0,I_d) 
\end{align*}

By replacing $\nabla f$ with its polar factorization, the procedure becomes:
$$  x^{(k+1)} = x^{(k)} - \gamma \nabla u \circ M(x^{(k)}) + \sqrt{2 \gamma} z^{(k)}$$
By studying $y^{(k)} = M(x^{(k)})$, one can see that the LMC procedure implies doing a preconditioned LMC algorithm on the convex function $u$.
\begin{align*}
   &y^{(k+1)} = M(x^{(k+1)}) \\
   &= M \left(x^{(k)} - \gamma \nabla u(y^{(k)}) + \sqrt{2 \gamma} z^{(k)}\right) \\
   &= M(x^{(k)}) + J_M(x^{(k)})\left[ - \gamma \nabla u(y^{(k)}) + \sqrt{2 \gamma} z^{(k)}\right] \\
   &\;+ \circ(\|\varepsilon\|) \\
   &= y^{(k)} - \gamma J_M(x^{(k)}) \nabla u(y^{(k)}) + \sqrt{2 \gamma} J_M(x^{(k)}) z^{(k)} \\ &\;+ \circ(\|\varepsilon\|)
\end{align*}
with $\varepsilon = - \gamma \nabla u(y^{(k)}) + \sqrt{2 \gamma} z^{(k)}$.
One can note that the preconditioned matrix $H = J_M(x^{(k)})$ is not necessarily positive definite.

\section{Augmented Bridge Matching}
Given a coupling $\Pi_{0,1}$ and random variables $(X_0, X_1)$, the augmented bridge matching algorithm \citep{debortoli2023augmented} aims at learning a stochastic dynamic mapping between $X_0$ and $X_1$ that preserves the coupling $\Pi_{0,1}$. 

In the probability space of path measures $\cali{P}(\cali{C}([0,1], \R^d))$,
let $\cali{M}$ denotes the path measures associated to the SDE $dX_t = v_t(X_t)\text{d}t + \sigma_t \text{d}B_t$, the functions $\sigma$ and $v$ being locally Lipschitz. Given a path measure $\mathbb{Q} \in \cali{M}$, the diffusion bridge of $\mathbb{Q}$ which is the distribution of $\mathbb{Q}$ conditioned on both endpoint is denoted by $\mathbb{Q}_{|0,1}$. The set of path measures considered to bridge $\cali{P}(X_0)$ and  $\cali{P}(X_1)$ according to the coupling $\Pi_{0,1}$ is $\Pi_{0,1} \mathbb{Q}_{|0,1} = \int_{\R^d \times \R^d} \mathbb{Q}_{|0,1}(.|x_0, x_1) \Pi_{0,1}(\text{d}x_0,\text{d}x_1)$. In \citet{debortoli2023augmented}, the authors showed that under mild conditions,  $\Pi_{0,1} \mathbb{Q}_{|0,1}$ was associated to the following SDE:
$$\text{d}X_t = \{b_t(X_t) + \sigma_t^2 u_t\} \text{d}t + \sigma_t \text{d}B_t, \qquad X_0 \sim \mu$$

with $u_t = \E_{\mathbb{P}_{1|0,t}} \left[\nabla \log \mathbb{Q}_{1|t}(X_1 | X_t) | X_0, X_t  \right]$ where $\mathbb{Q}_{1|t}$ and $\mathbb{P}_{1|0,t}$ are respectively the conditional distribution of $\mathbb{Q}$ at time $1$ given the state at time $t$ and the conditional distribution of $\mathbb{P}$ at time $1$ given the coupling state at time $0$ and $t$.

This SDE gives a way to sample from $\Pi_{0,1}$ by first sampling $X_0 \sim \mu$ and then discretize the SDE to get $X_1$. Because $u_t$ is intractable, it is approximated by a neural network $u_t^{\theta}$ learned to minimize the regression loss: 
$$\int_0^1 \lambda_t \E[\|u_t^{\theta}(X_0, X_t) - \nabla \log \mathbb{Q}_{1|t}(X_1 | X_t)\|^2] \text{d}\mathbb{P}(X_0, X_t, X_1)$$

A particular case of diffusion bridge is the Brownian bridge $\mathbb{Q}_{|0,1}$ for which $v = 0$ and $\sigma_t = \sigma$ which is the one usually used in practice. 

\section{ICNNs Benchmark}
We compare three different architectures: Linear which is the one proposed in \citet{amos2023amortizing}, FQuad with the same architecture but with a extra full quadratic layer at the end and ours (with a rank of $1$ for the intermediate quadratic layers). Given input dimension $d$, all ICNNs have $4$ layers, with hidden states $z_{i}$ of size $[d,d,d,d]$ for our method, $[2d, 2d, 2d, 2d]$ for Linear, FQuad so that architectures have a comparable number of parameters. The ICNN optimization is done with a constant learning rate of $0.0005$ for $d=32$ and $d=128$. The networks have been trained for $60000$ iterations. In all experiments, the distances have been computed with batches of size $4096$. In the first experiment, the starting measure $\mu$ is a $d$-dimensional standard Gaussian, the target $\nu$ is a gaussian of mean $0$ and covariance matrix $\text{diag}(1, 2, ..., d)$. Because both source and target distribution are gaussians, the OT map is known in closed form~\citet{peyré2020computational} and the $\mathcal{L}_2$ unexplained variance percentage \citep{makkuva2020optimal, korotin2021neural} is used to assess the accuracy of the computed OT map. The estimation of this metric is done using batches of size $2048$. In the second experiment, the starting measure $\mu$ is still a $d$-dimensional standard Gaussian and the objective is to map multimodal data whose modes have various sizes. More precisely, the target distribution is composed of 7 gaussians whose means are respectively : $(30,0,30,0,30,0,...)$ $(-30,0,-30,0,-30,0,...)$ $(0,-30,0,-30,0,-30,...)$ $(\frac{30}{\sqrt{2}}, \frac{30}{\sqrt{2}}, \frac{30}{\sqrt{2}}, \frac{30}{\sqrt{2}}, ...)$ $(\frac{30}{\sqrt{2}}, \frac{-30}{\sqrt{2}}, \frac{30}{\sqrt{2}}, \frac{-30}{\sqrt{2}}, ...)$ $(\frac{-30}{\sqrt{2}}, \frac{30}{\sqrt{2}}, \frac{-30}{\sqrt{2}}, \frac{30}{\sqrt{2}}, ...)$ and their covariance matrices are : $I_d$, $2 I_d$, $3 I_d$, $4 I_d$, $5 I_d$, $6 I_d$, $7 I_d$. This time, the OT map is not known in closed form and we only use the Sinkhorn divergence to estimate the distance between the generated distribution $\nabla u_{\theta} {}_\# \mu$ and the target $\nu$ in order to compare the three architectures. 

\begin{table}[h!]
 \centering
    \begin{tabular}{|c|c|c|}
\hline
dimension $d$ & $32$ & $128$ \\
\hline
Linear    &$20\,577$   & $328\,065$ \\
FQuad     & $20\,673$   & $328\,449$    \\
Ours    & $15\,714$   & $247\,170$ \\
\hline
\end{tabular}
\caption{Number of parameters for the 3 architectures under consideration.}
    \label{tab_params}
\end{table}

\begin{table}[h!]
 \centering
    \begin{tabular}{|l|c|c|c|}
\hline
dimension $d$ &  $32$ & $128$\\
\hline
Linear    &$0.72 \color{gray}\pm 0.05$   &$1.26 \color{gray}\pm 0.08$ \\
FQuad     &$0.38 \color{gray}\pm 0.03$   &$0.97 \color{gray}\pm 0.06$ \\
Ours      & $\mathbf{0.031 \color{gray}\pm 0.003}$   & $\mathbf{0.082 \color{gray}\pm 0.002}$ \\
\hline
\end{tabular}
\caption{$\mathcal{L}_2$ unexplained variance percentage for the 3 architectures considered when the target is the multivariate gaussian.}
    \label{tab_icnn_bench_gauss}
\end{table}

\begin{table}[h!]
 \centering
    \begin{tabular}{|c|c|c|c|c|c|}
\hline
dimension $d$ & $32$ & $128$ \\
\hline
Linear   & $548 \color{gray}\pm 75$   & $246 \times 1e^1 \color{gray}\pm 42\times 1e^1$ \\
FQuad   &  $452 \color{gray}\pm 77$   & $176 \times 1e^1 \color{gray}\pm 14\times 1e^1$ \\
Ours  & $\mathbf{234 \color{gray}\pm 42}$   & $\mathbf{104 \times 1e^1 \color{gray}\pm 25\times 1e^1}$ \\
\hline
\end{tabular}
\caption{Sinkhorn divergence results for the 3 architectures considered when the target is the mixture of gaussians.}
    \label{tab_icnn_bench_multi}
\end{table}

\section{Topography Experiments}
\subsection{Creation of the dataset}
We used the Python package $\texttt{elevation}$ to get the elevation of three different regions of the globe: Chamonix, London, and Cyprus. Given the latitudes and longitudes of the desired area, $\texttt{elevation}$ returns a grid of the area with the elevation value at each grid point. For the Chamonix example, we obtained $323932$ points $(x,y) \in \R^2$ and their corresponding elevation. We dequantized the elevations by adding a uniform noise on [0,1] to them before using a Gaussian filter to make the gradients smoother. To do this, we used the function $\texttt{gaussian\_filter}$ from the $\texttt{scipy}$ library. We then numerically estimate the gradients associated with the elevation and obtain a dataset of $323932$ points in $\R^2$ and the associated gradients in $\R^2$ for the example of Chamonix. We obtained data for the Cyprus and London regions in the same way. 

\subsection{London and Cyprus results}

\begin{figure*}[h!]
    \centering
    \begin{tabular}{cccc}

\includegraphics[scale=0.055, trim = 0cm 0cm 0cm 0cm, clip]{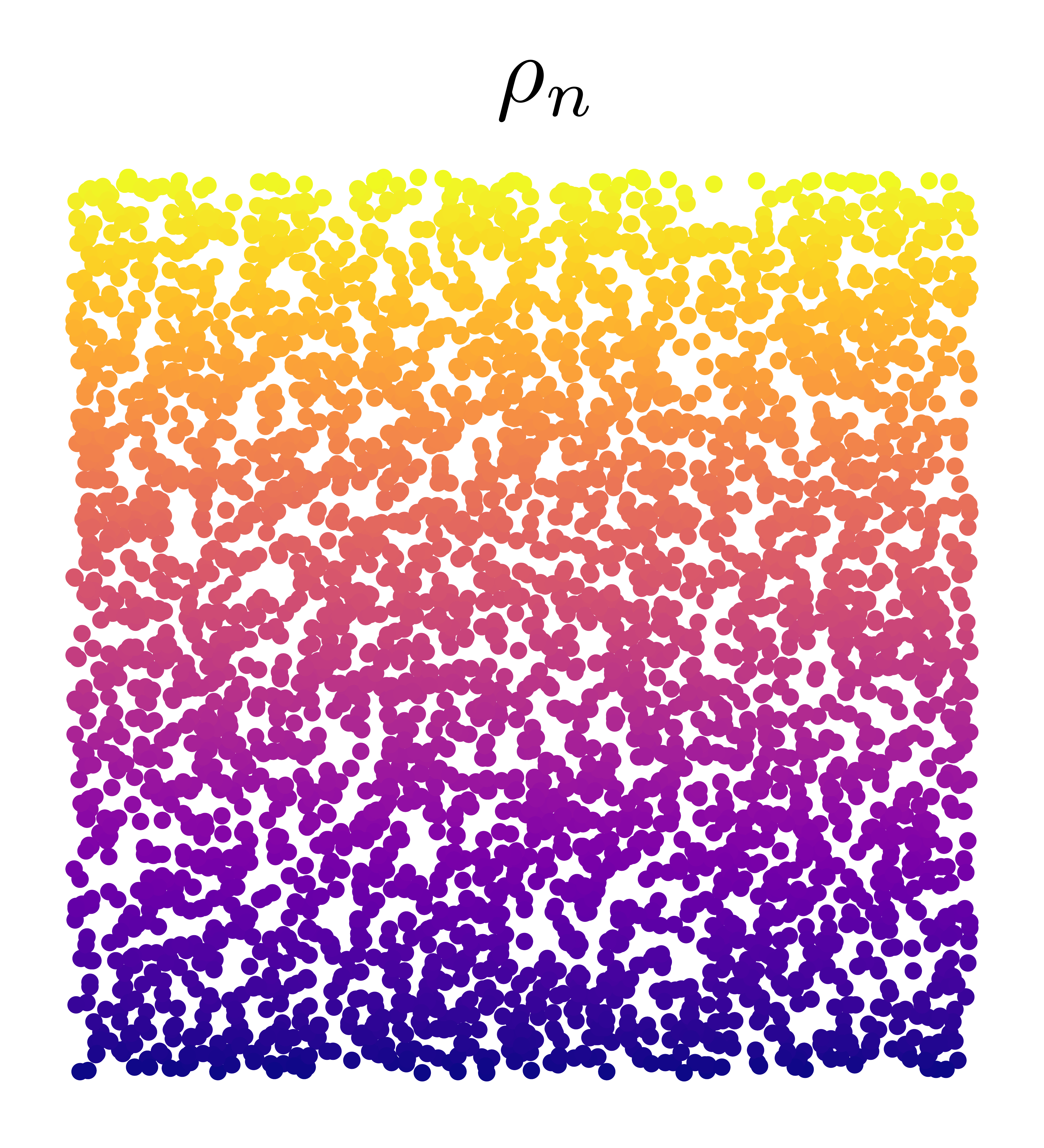}
\includegraphics[scale=0.055, trim = 0cm 0cm 0cm 0cm, clip]{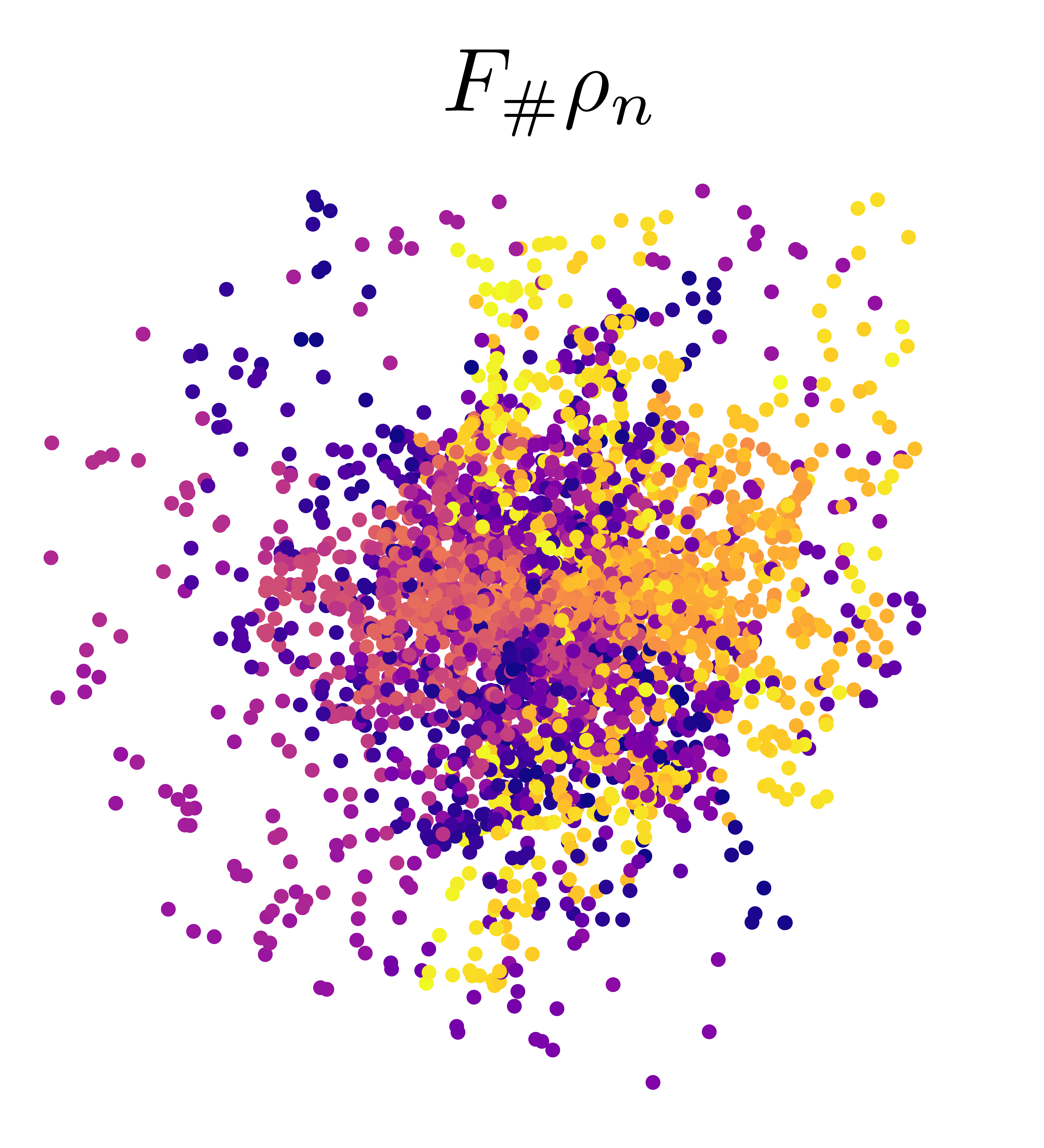} &
\includegraphics[scale=0.055, trim = 0cm 0cm 0cm 0cm, clip]{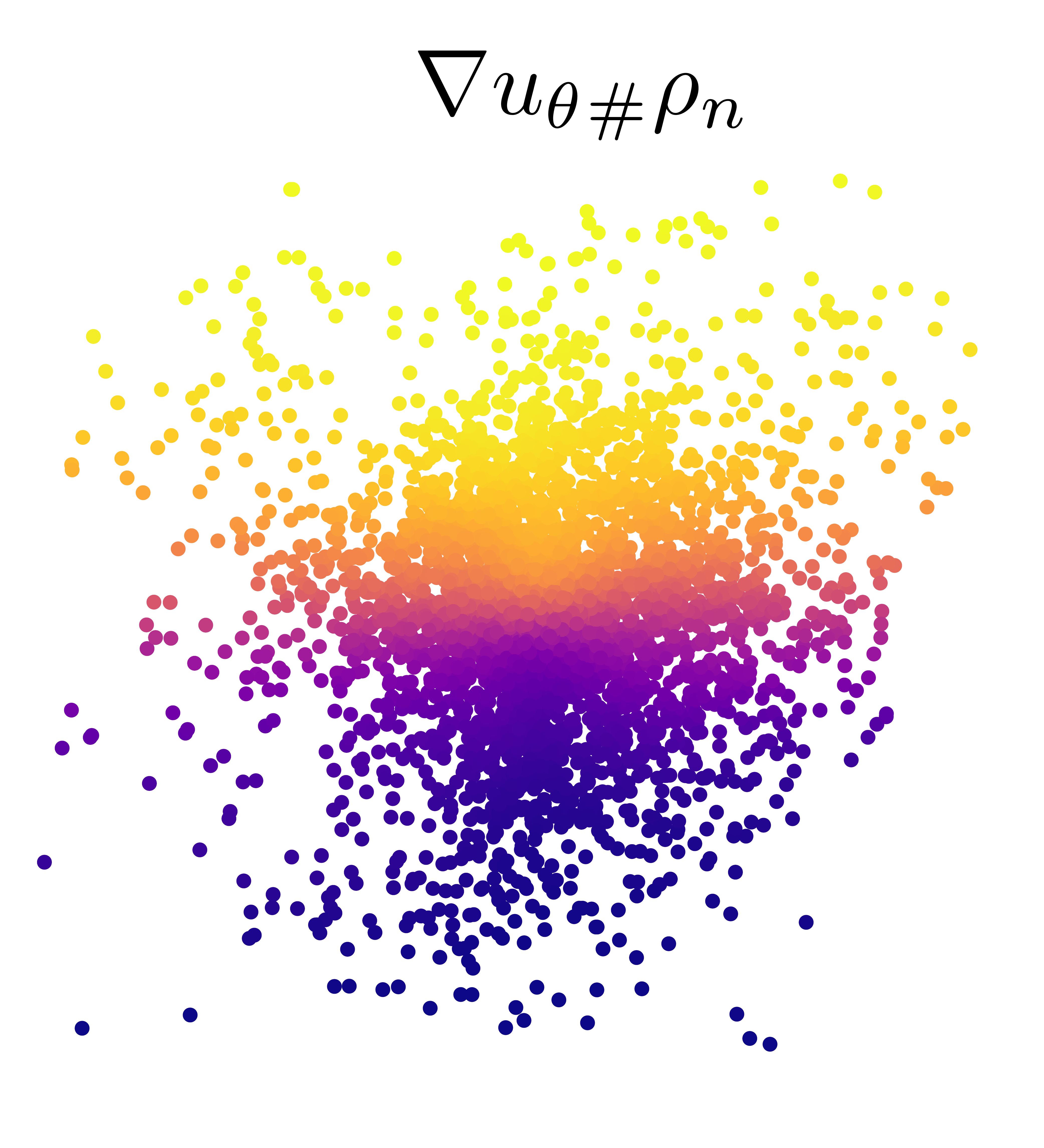} &
\includegraphics[scale=0.055, trim = 0cm 0cm 0cm 0cm, clip]{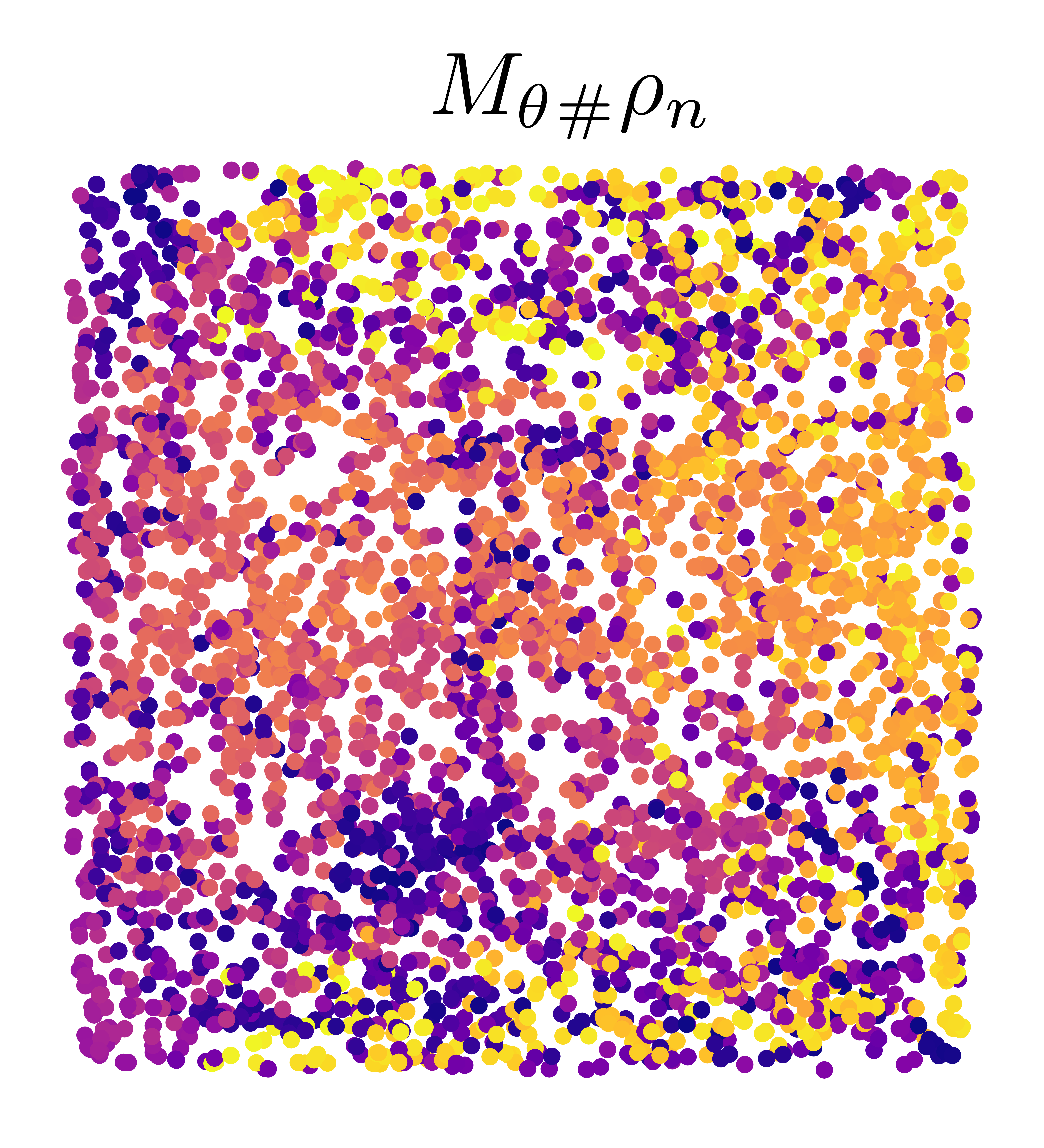} \\
\end{tabular}
    \caption{Respective actions of the learned vector fields associated with the polar factorization of the elevation gradient in the London area.}
    \label{fig:topo Londres}
\end{figure*}

\begin{figure}[h!]
    \centering
    \begin{tabular}{c}

\includegraphics[scale=0.12, trim = 0cm 25cm 0cm 25cm, clip]{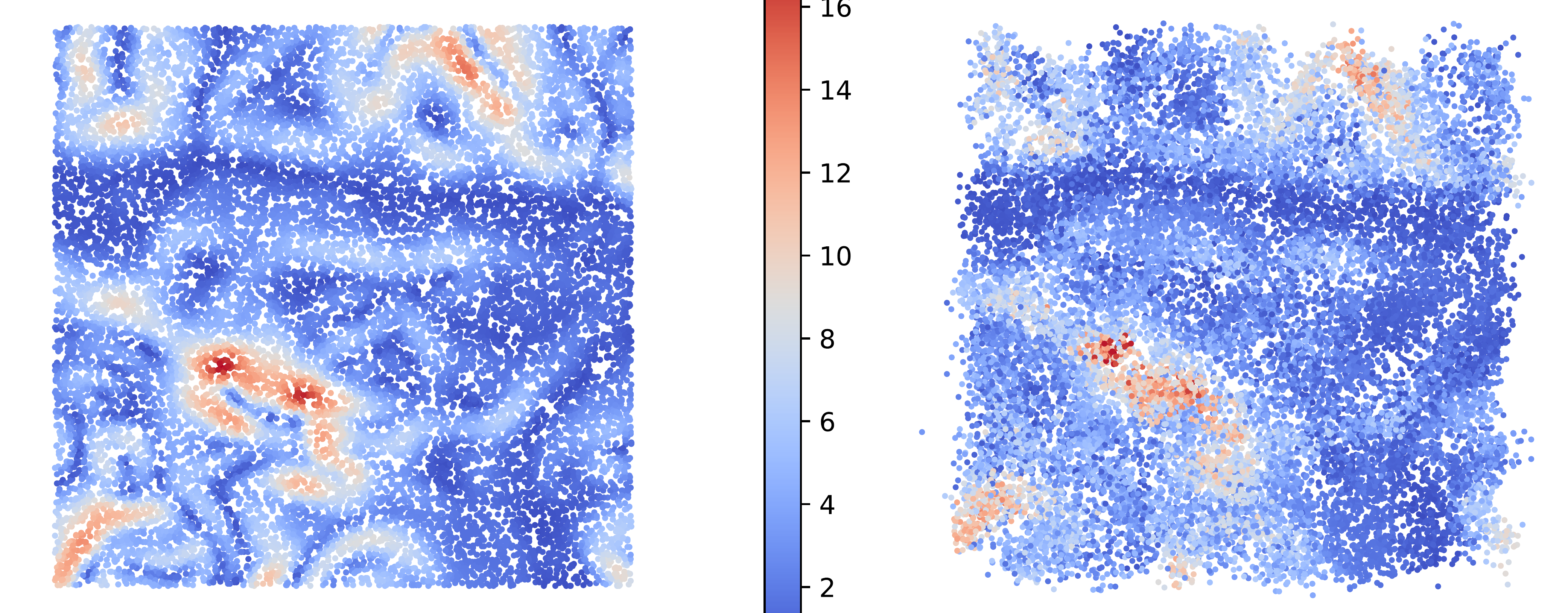}
\end{tabular}
    \caption{$I_{\psi}$'s ability to replace gradients in the original $\Omega$ space for the example of London region's elevation gradient. }
    \label{fig:topo Londres}
\end{figure}

\begin{figure*}[h!]
    \centering
    \begin{tabular}{cccc}

\includegraphics[scale=0.055, trim = 0cm 0cm 0cm 0cm, clip]{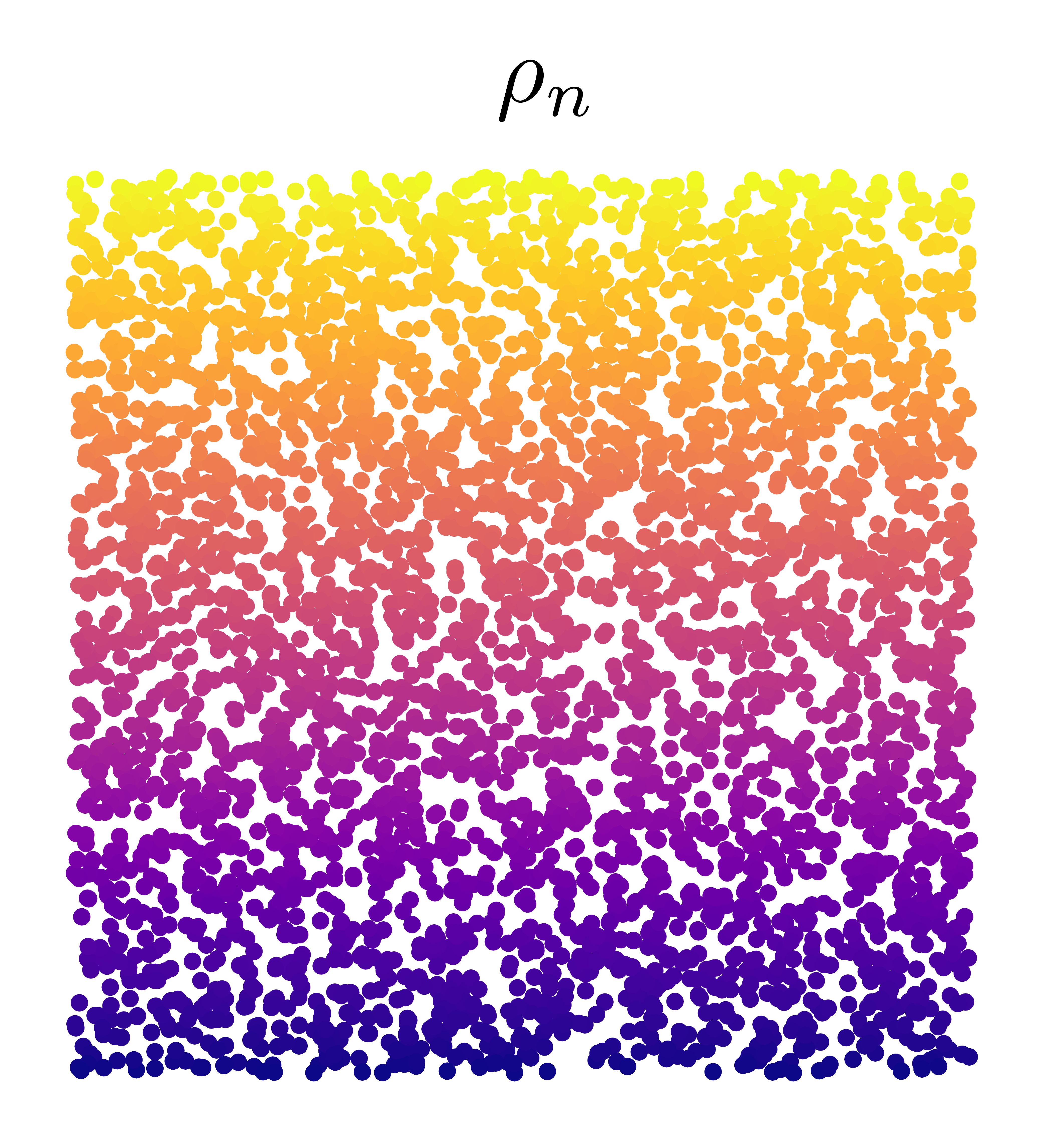}
\includegraphics[scale=0.055, trim = 0cm 0cm 0cm 0cm, clip]{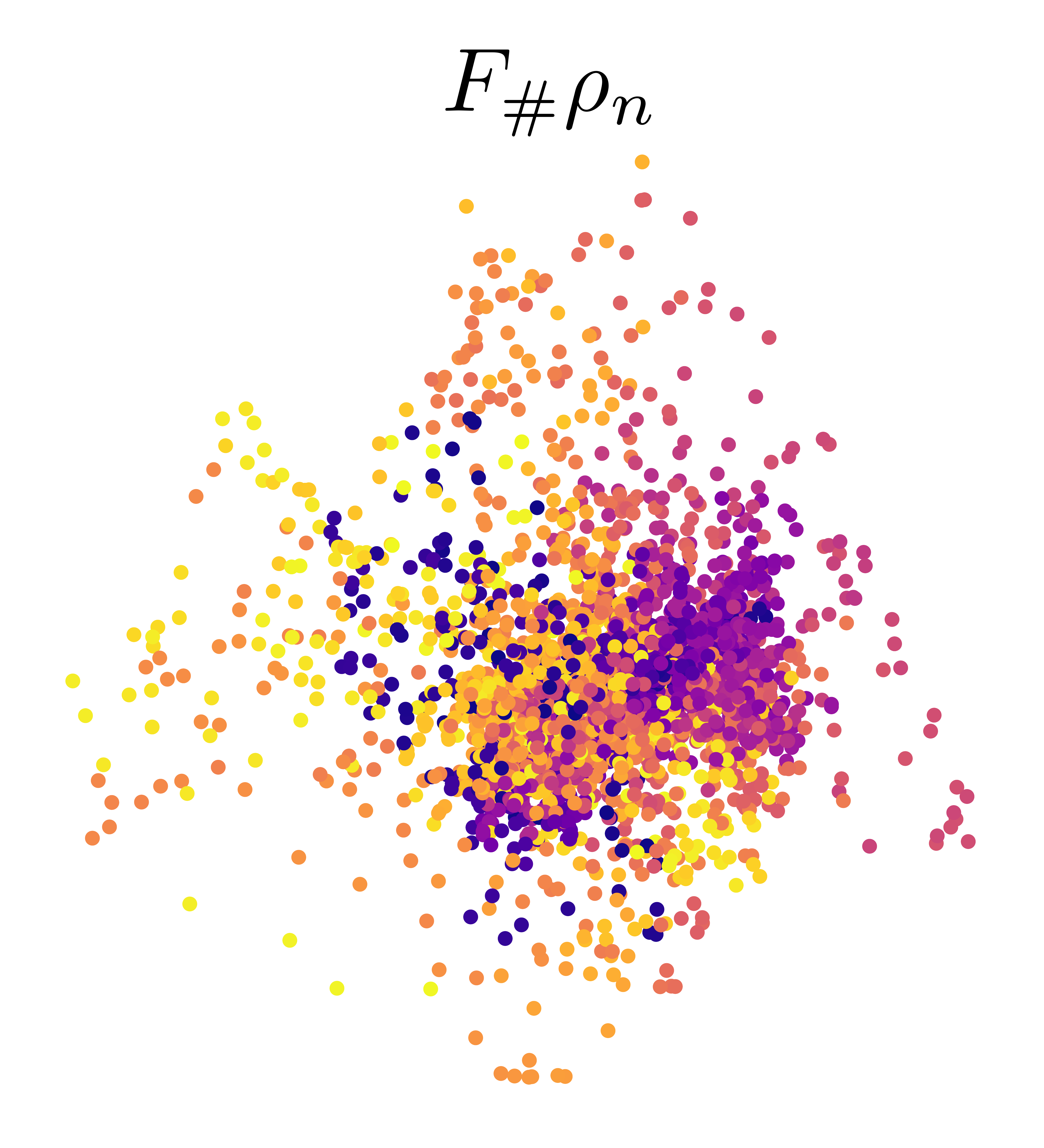} &
\includegraphics[scale=0.055, trim = 0cm 0cm 0cm 0cm, clip]{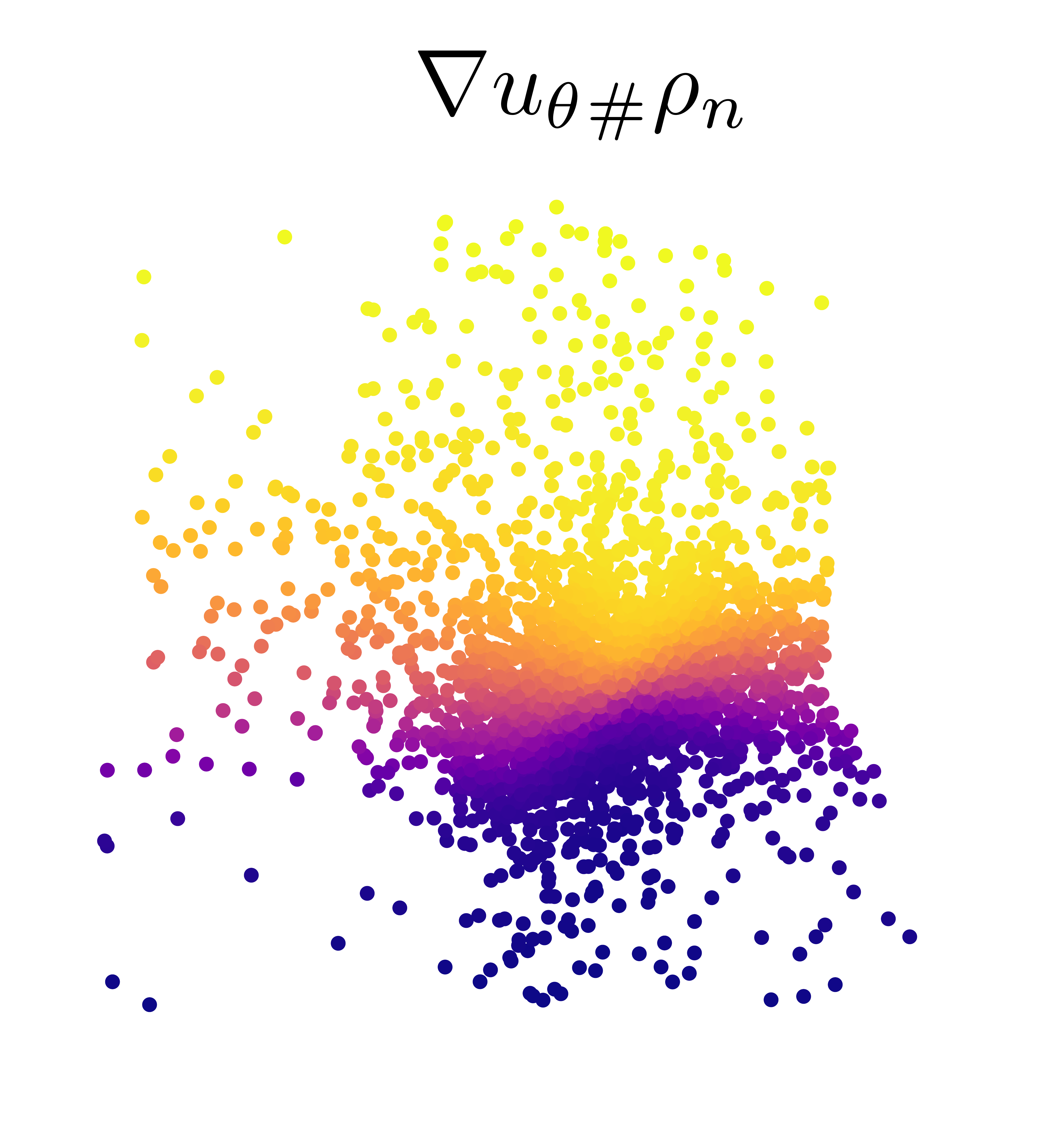} &
\includegraphics[scale=0.055, trim = 0cm 0cm 0cm 0cm, clip]{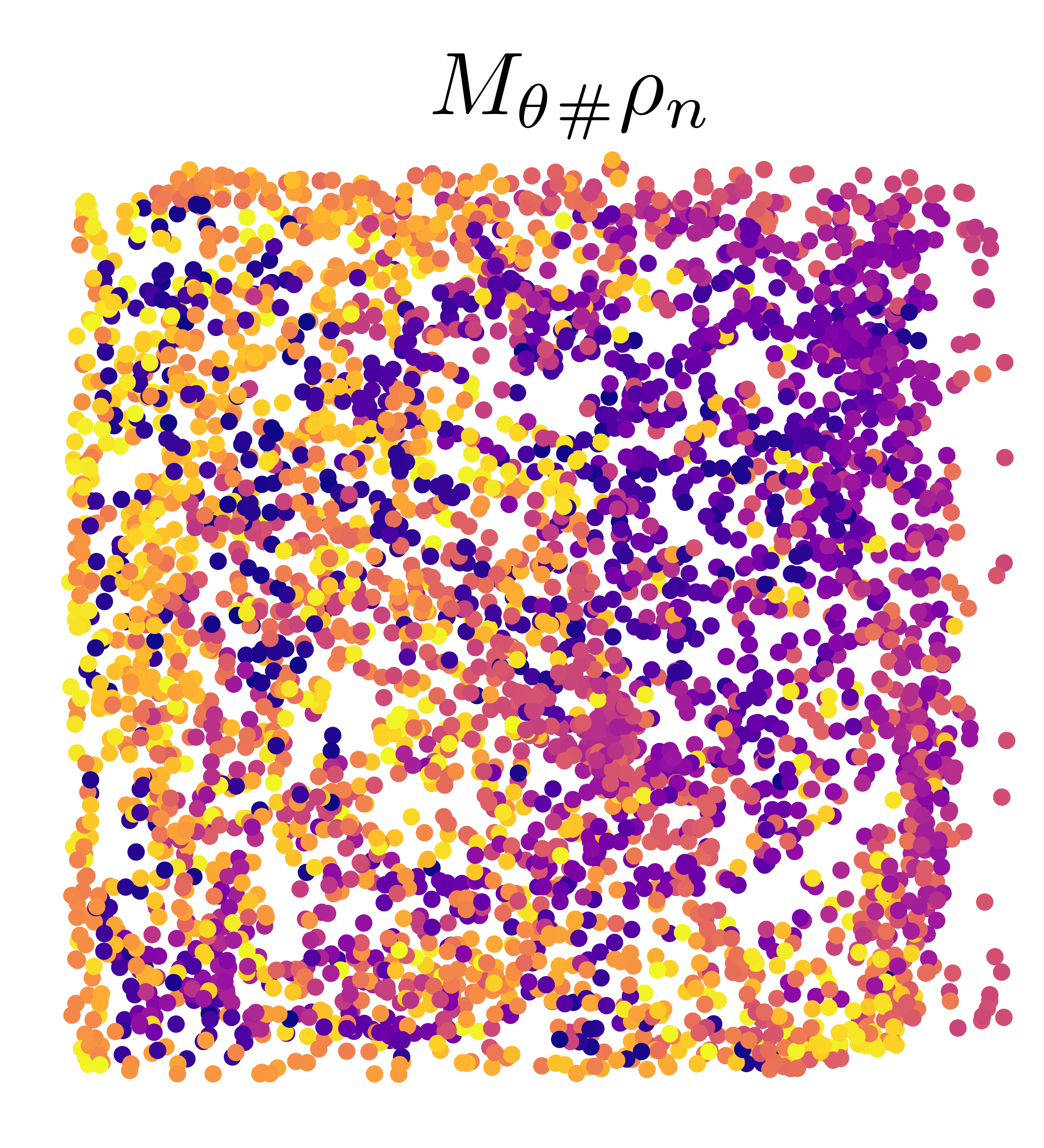} \\
\end{tabular}
    \caption{Respective actions of the learned vector fields associated with the polar factorization of the elevation gradient in the Cyprus neighborhood.}
    \label{fig:topo Chypre}
\end{figure*}

\begin{figure}[h!]
    \centering
    \begin{tabular}{c}

\includegraphics[scale=0.12, trim = 0cm 25cm 0cm 25cm, clip]{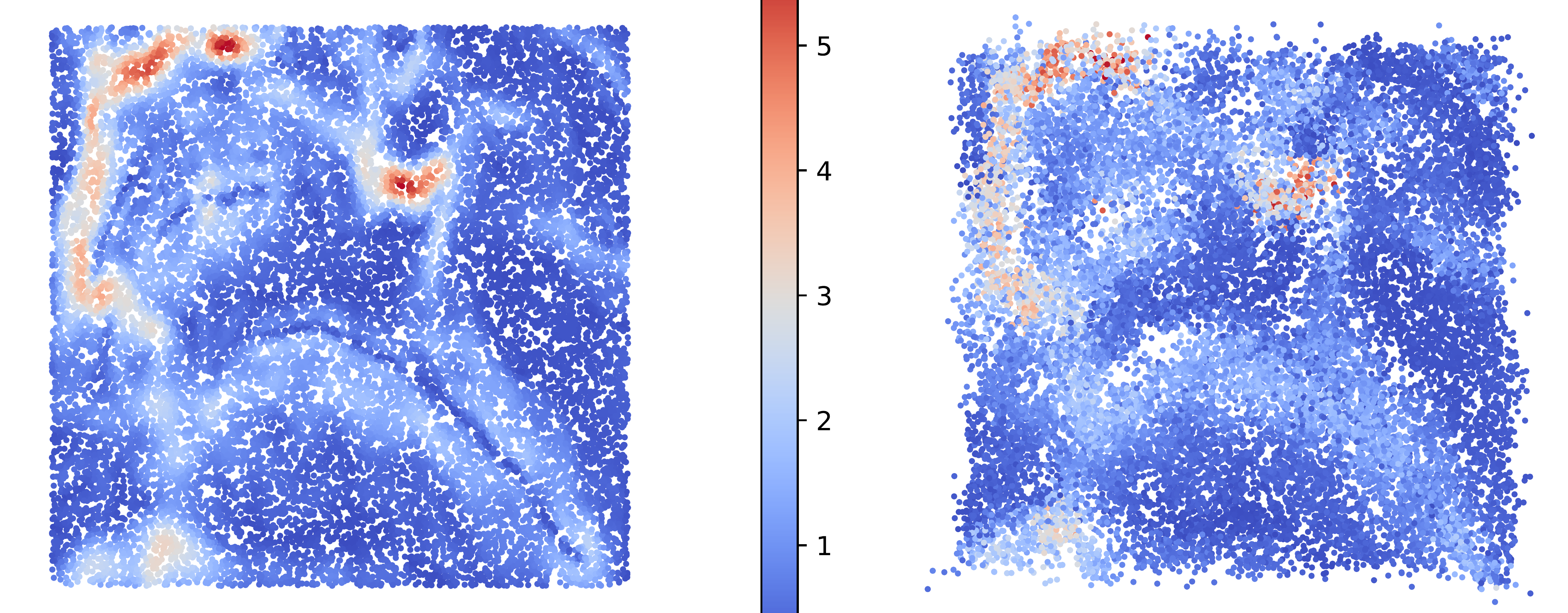}
\end{tabular}
    \caption{$I_{\psi}$'s ability to replace gradients in the original $\Omega$ space for the example of Cyprus region's elevation gradient. }
    \label{fig:topo Chypre}
\end{figure}

\begin{figure*}[h!]
    \centering
    \begin{tabular}{ |p{5.8cm}||p{1.4cm}|p{1.4cm}|p{1.4cm}|  }
 \hline
  $ $ & Chamonix  & London & Chypre\\
 \hline
$W_{\varepsilon}$($\nabla u_{\theta}{}_{\#} \rho_n$, $F_{\#} \rho_n$) & 0.36    &0.26&   0.021\\
 $W_{\varepsilon}$($\nabla F_{\#} \rho_n$, $F_{\#} \rho_n'$) &   0.31  & 0.25  &0.020\\
 $\E_{k, \mathbf{z}} \left[S_{\varepsilon}((I_{\psi}(M_{\theta}(\mathcal{B}_{\alpha}(x_k)), \mathbf{z}),\mathcal{B}_{\alpha}(x_k))\right ]$  &0.046 & 0.054&  0.038\\
 $S_{\varepsilon}$( $\rho_{128}$, $\rho_{128}'$) & 0.039 & 0.035 & 0.021 \\
 \hline
\end{tabular}
    \caption{NPF and $I_{\psi}$ performances for the topography experiments, $n=2048$.}
    \label{fig:tab_topo_bis}
\end{figure*}

\FloatBarrier

\section{LeNet classifier Experiments}
\subsection{LeNet classifier architecture}
\begin{figure}[h!]
    \centering
        \input{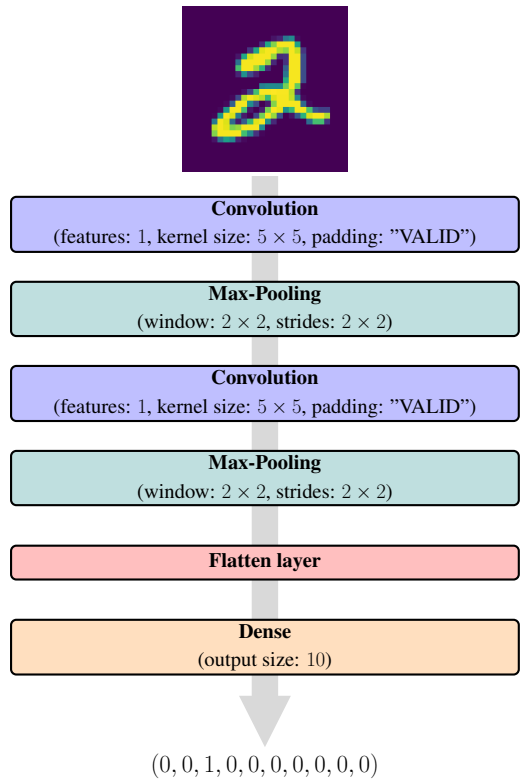} 
    \caption{Le Net classifier architecture used in experiments.}
    \label{fig:le_net_arch}
\end{figure}

The LeNet classifier architecture used for the experiments is composed of two convolutive layers followed by a relu activation function and a max pooling; it ends with a dense layer as described in \Cref{fig:le_net_arch}.
The optimization space that we consider is $\Omega = [-1, 1]^{222}$.

\subsection{Complementary graphs for the MNIST classifier experiments}
\begin{figure*}[h!]
    \centering
    \begin{tabular}{ccc}
\includegraphics[scale=0.28, trim = 0cm 0cm 0cm 0cm, clip]{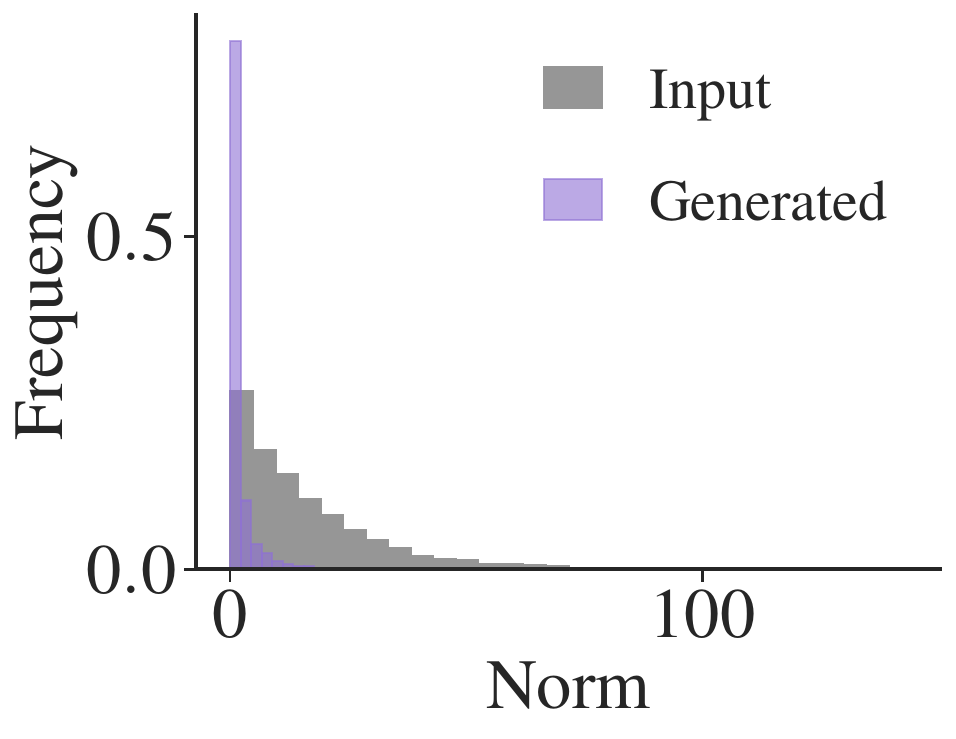} & 
\includegraphics[scale=0.28, trim = 0cm 0cm 0cm 0cm, clip]{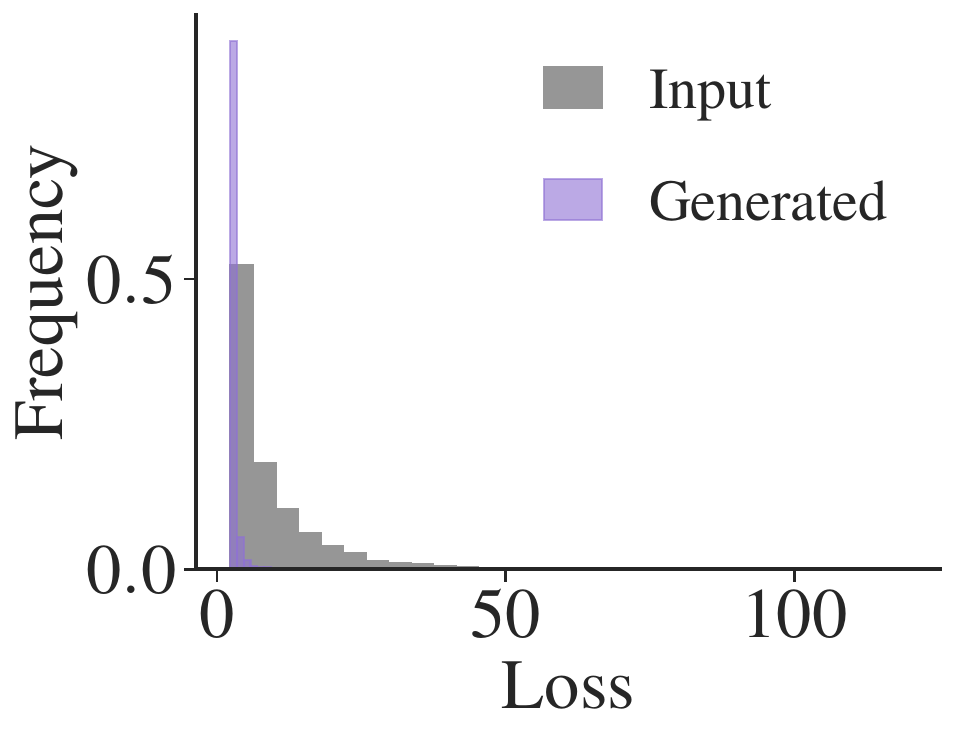} &
\includegraphics[scale=0.28, trim = 0cm 0cm 0cm 0cm, clip]{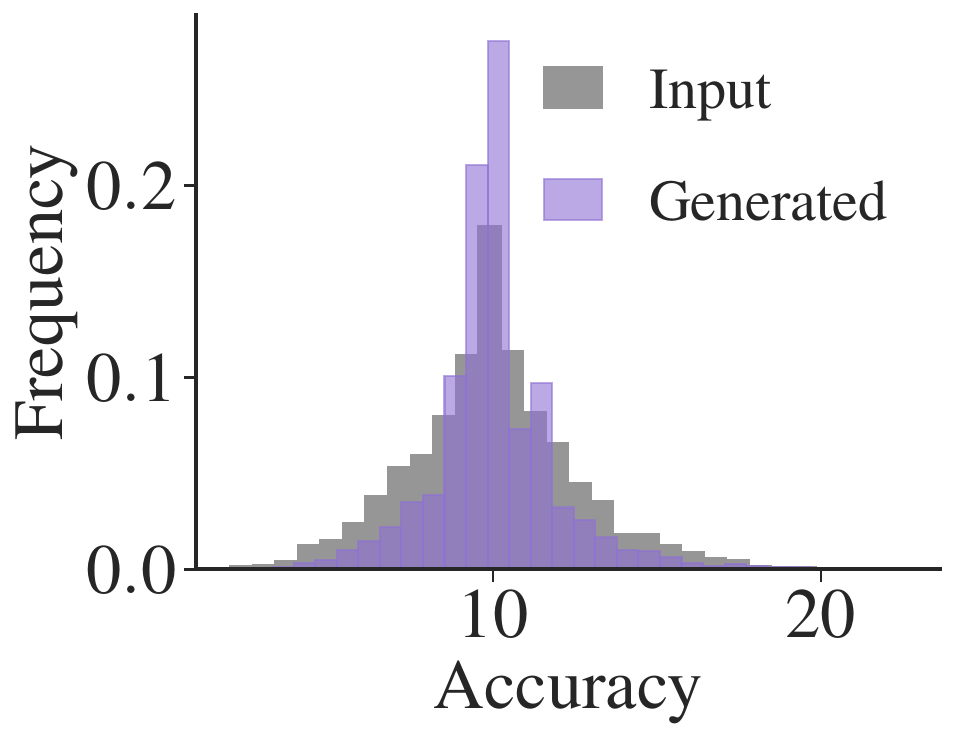}
\end{tabular}
    \caption{Characteristics of the generated critical points for the \S\ref{subsec:npf} MNIST classifier experiment. In gray the classifier weights have been drawn uniformly in the $\Omega$ optimization space, while in purple the weights have been drawn using $I_{\psi}(\nabla u_{\theta}^*(0), \mathbf{z})$.}
    \label{fig:graph_all}
\end{figure*}

\begin{figure*}[h!]
    \centering
    \begin{tabular}{ccc}
\includegraphics[scale=0.28, trim = 0cm 0cm 0cm 0cm, clip]{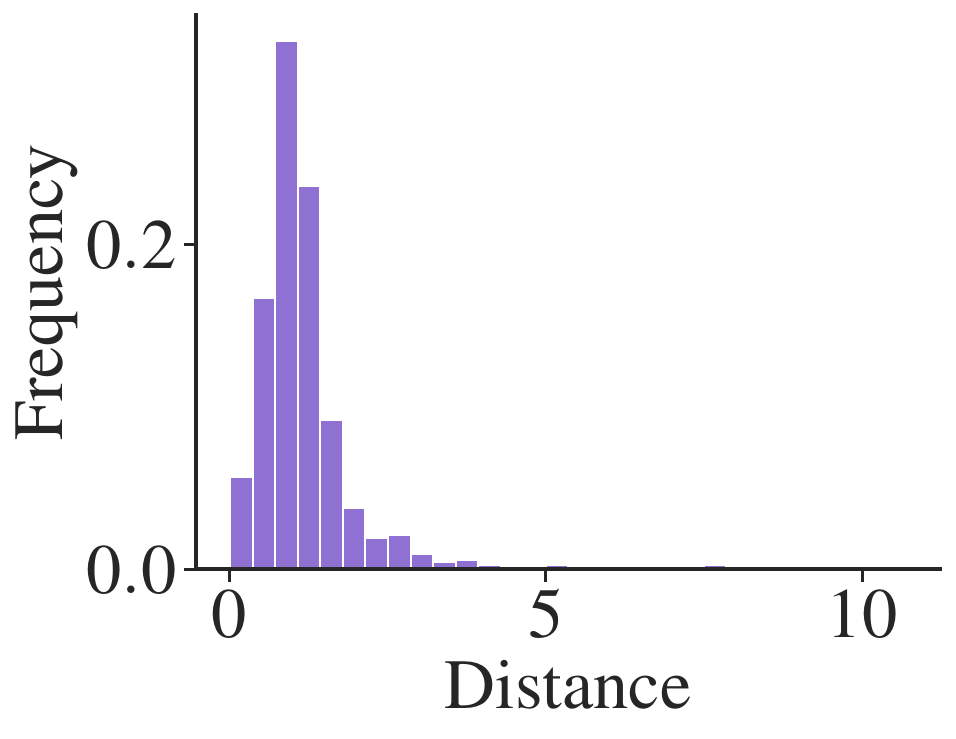} & 
\includegraphics[scale=0.28, trim = 0cm 0cm 0cm 0cm, clip]{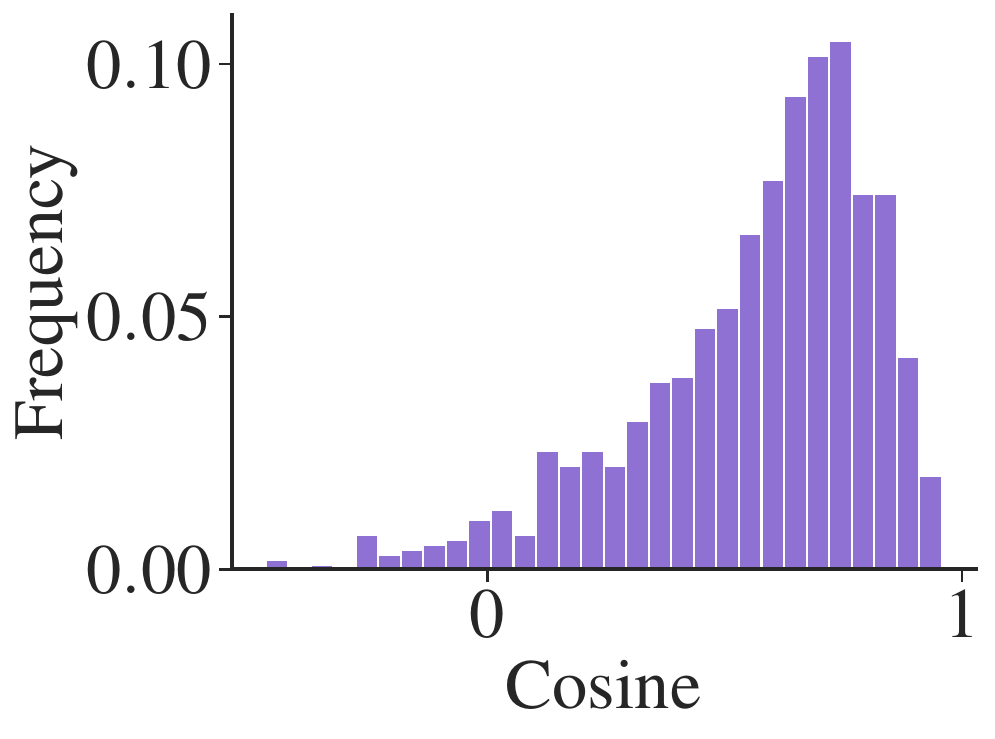} &
\includegraphics[scale=0.28, trim = 0cm 0cm 0cm 0cm, clip]{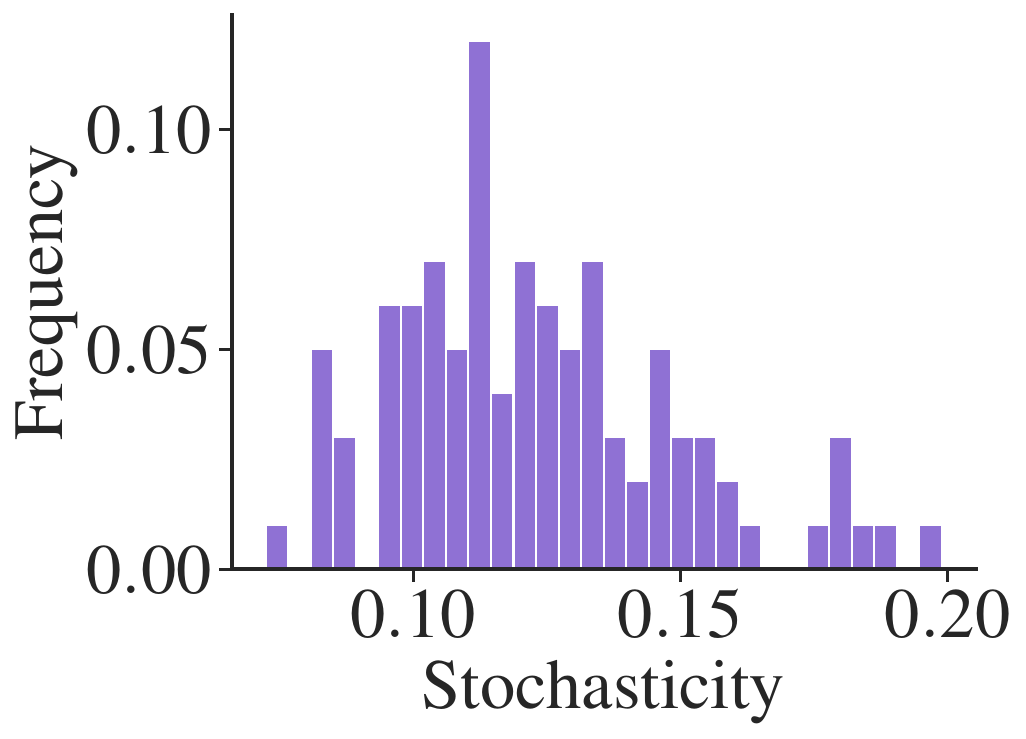} 
\end{tabular}
    \caption{Performances of $I_{\psi}$ for the \S\ref{subsec:exp-mnist} MNIST classifier experiment and stochasticity of the MNIST loss. The distance and cosine plots demonstrate the ability of $I_{\psi}$ to correctly generate weights with a fixed gradient. For the stochasticity plot, the classifier weights are fixed to the weights obtained after gradient descent and different minibatches of MNIST images are used to compute the gradient of the loss function. The stochasticity plot shows the distribution of the sinkhorn divergence between two gradient batches computed from the same weights.}
    \label{fig:graph_desc_part}
\end{figure*}

\begin{figure*}[h!]
    \centering
    \begin{tabular}{ccc}
\includegraphics[scale=0.28, trim = 0cm 0cm 0cm 0cm, clip]{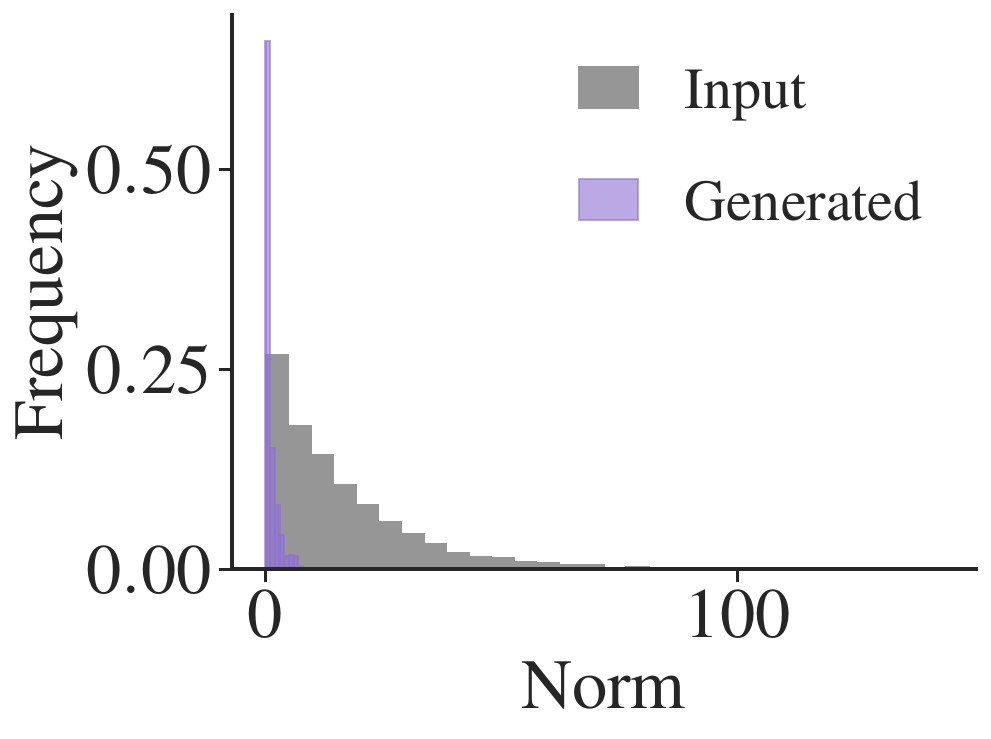} & 
\includegraphics[scale=0.28, trim = 0cm 0cm 0cm 0cm, clip]{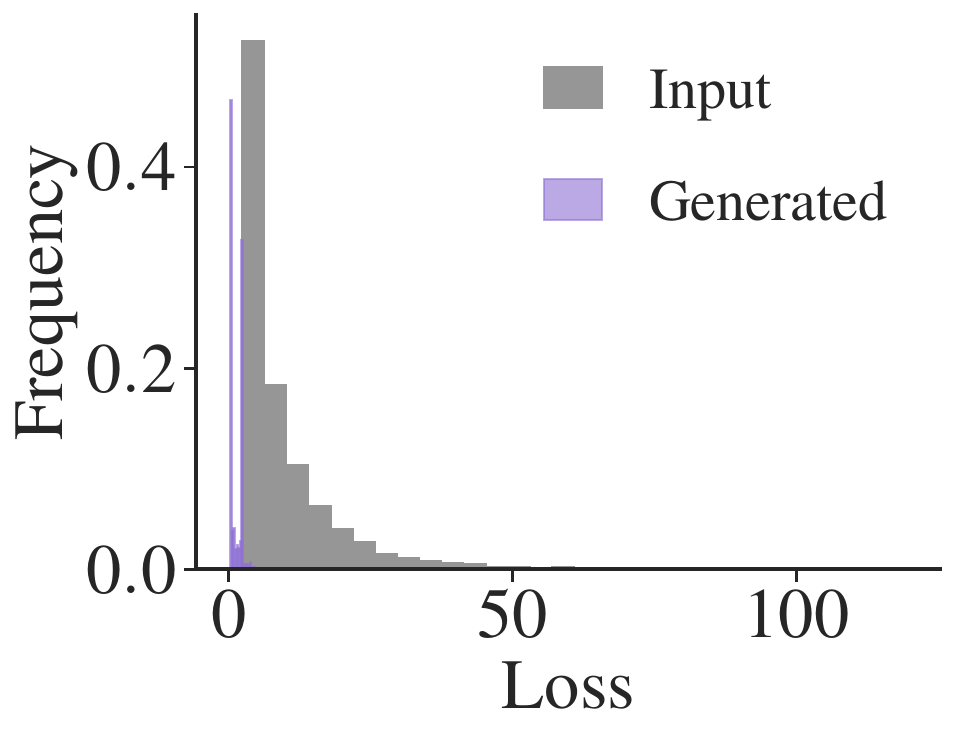} &
\includegraphics[scale=0.28, trim = 0cm 0cm 0cm 0cm, clip]{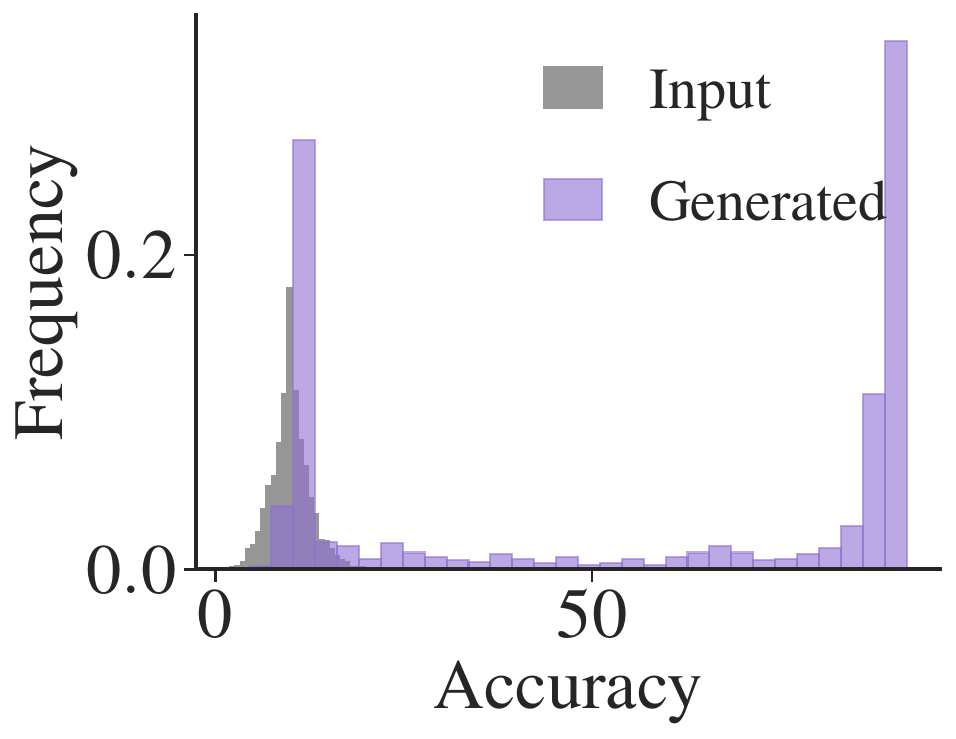}
\end{tabular}
    \caption{Characteristics of the generated critical points for the \S\ref{subsec:exp-mnist} MNIST classifier experiment. In gray the classifier weights have been drawn uniformly in the $\Omega$ optimization space, while in purple the weights have been drawn using $I_{\psi}(\nabla u_{\theta}^*(0), \mathbf{z})$.}
    \label{fig:graph_ccp}
\end{figure*}

\begin{figure*}[h!]
    \centering
    \begin{tabular}{ccc}
\includegraphics[scale=0.28, trim = 0cm 0cm 0cm 0cm, clip]{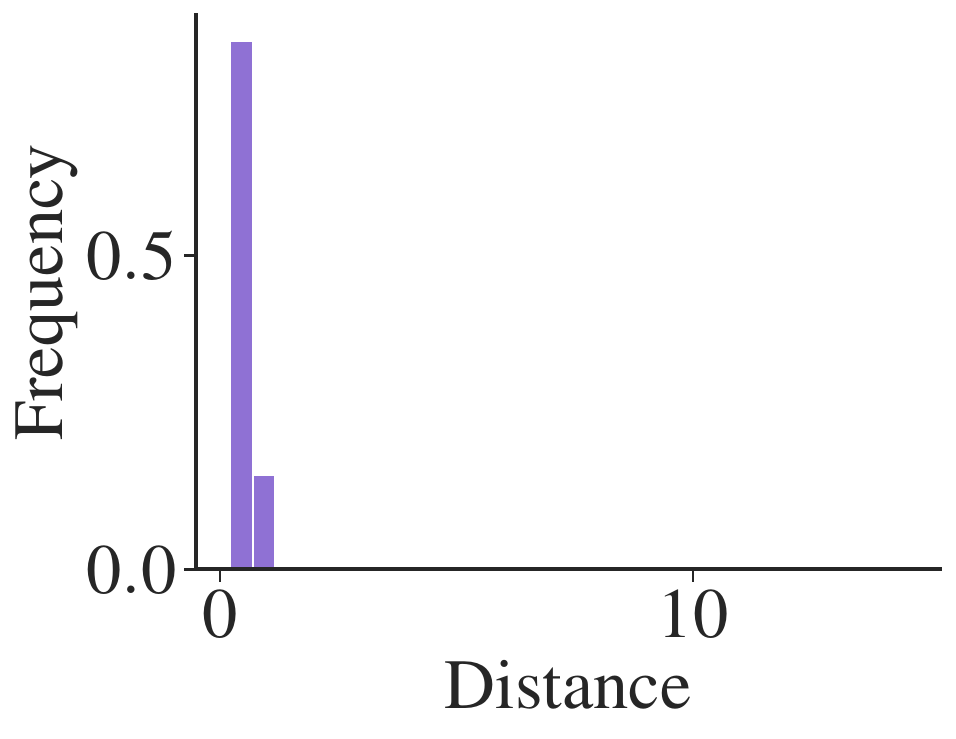} & 
\includegraphics[scale=0.28, trim = 0cm 0cm 0cm 0cm, clip]{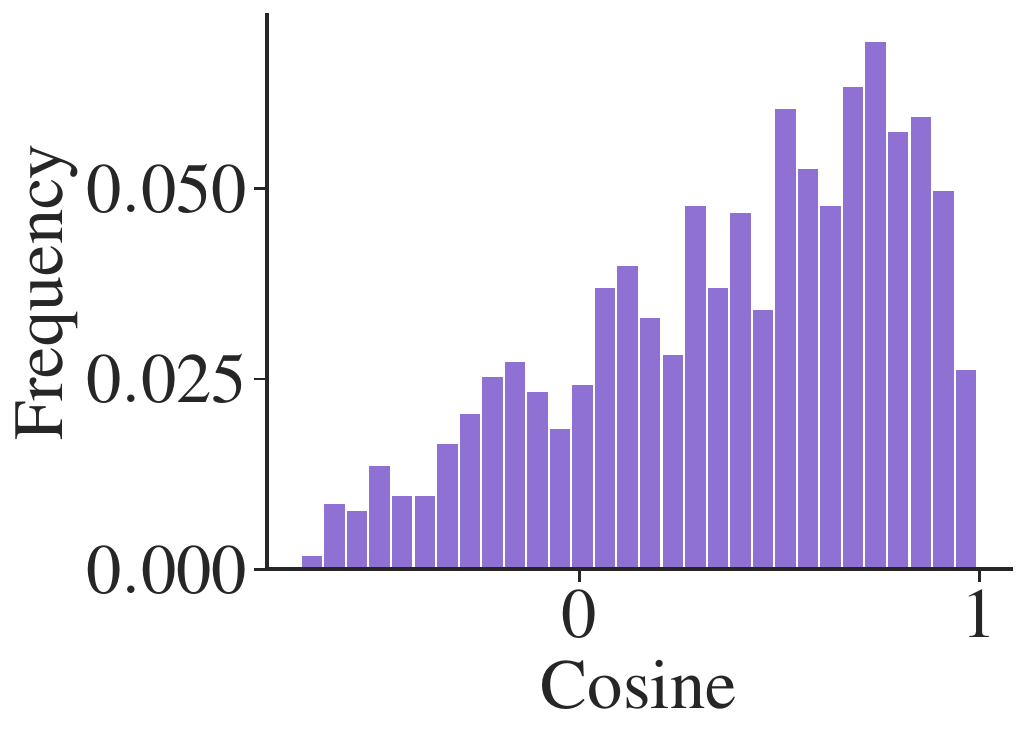} &
\includegraphics[scale=0.28, trim = 0cm 0cm 0cm 0cm, clip]{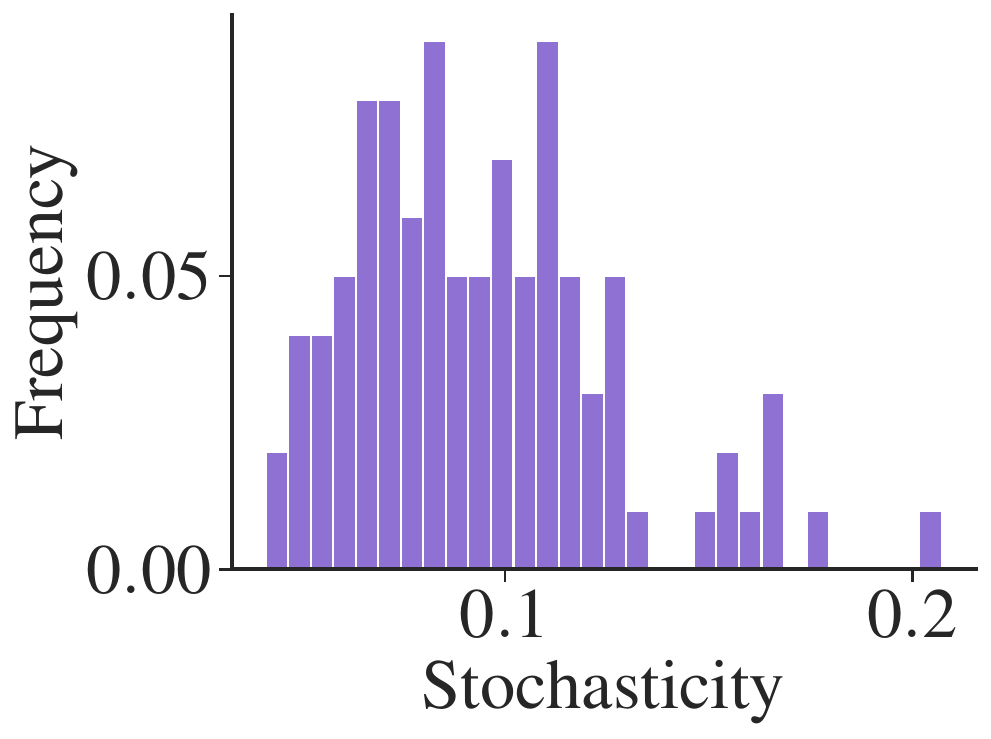} 
\end{tabular}
    \caption{Performances of $I_{\psi}$ for the \S\ref{subsec:lmc_npf} sampling MNIST classifier experiment. The distance and cosine plots demonstrate the ability of $I_{\psi}$ to correctly generate weights with a fixed gradient. For the stochasticity plot, the classifier weights are fixed to the final sampled particles and different minibatches of MNIST images are used to compute the gradient of the loss function. The stochasticity plot shows the distribution of the sinkhorn divergence between two gradient batches computed from the same weights.}
    \label{fig:graph}
\end{figure*}

\begin{figure}[h!]
    \centering
    \begin{tabular}{ccc}
\includegraphics[scale=0.28, trim = 0cm 0cm 0cm 0cm, clip]{sections/images/images_class_lenet_samp/norm_sampling.pdf} & 
\includegraphics[scale=0.28, trim = 0cm 0cm 0cm 0cm, clip]{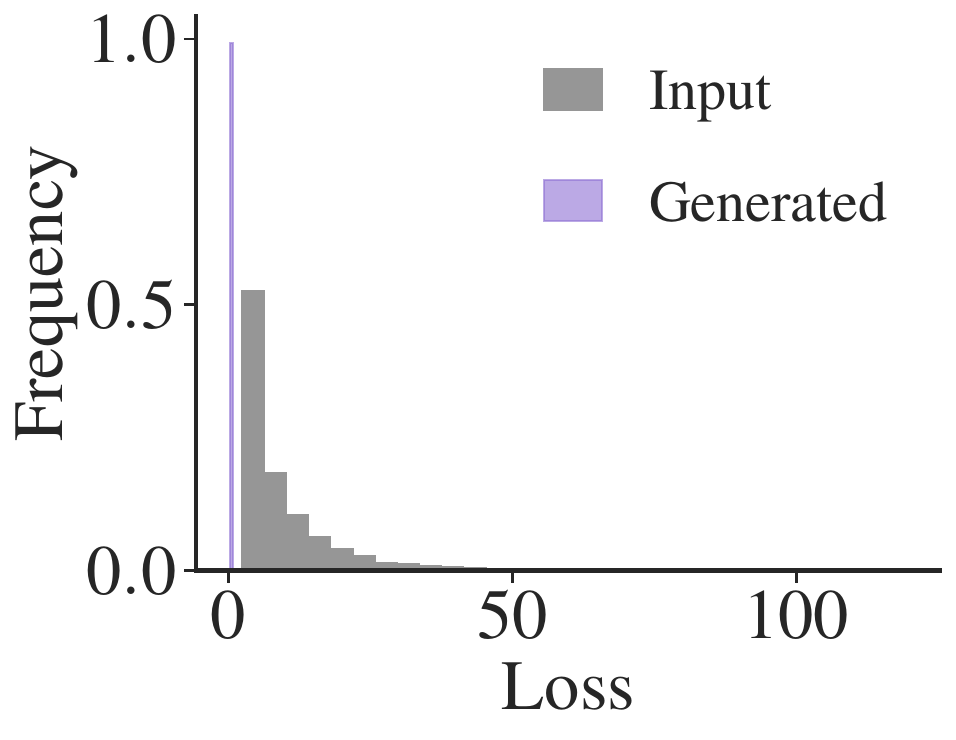} & 
\includegraphics[scale=0.28, trim = 0cm 0cm 0cm 0cm, clip]{sections/images/images_class_lenet_samp/accuracy_sampling.pdf}
\end{tabular}
    \caption{Characteristics of the points sampled using \Cref{alg:seq} for the \S\ref{subsec:lmc_npf} MNIST classifier experiment. In gray the classifier weights have been drawn uniformly in the $\Omega$ optimization space, while in purple the weights have been sampled using \Cref{alg:seq}. The generated samples are critical points which are good minima, as shown by the accuracy statistics.}
    \label{fig:graph_samp_crit}
\end{figure}

\FloatBarrier

\section{Hyperparameters}
\subsection{Parameterize $u_{\theta}$}
In all experiments, the convex function $u$ is parameterized using an ICNN $u_{\theta}$ whose architecture is detailed in \S\ref{subsec:icnn_arch}. The rank of the quadratic term $Q_{A,\delta}(x)$ is always taken equal to $1$, which means that $A$ is a row matrix. \textbf{We noted that it was necessary to choose smooth activation functions in $u$'s parameterization to avoid convergence problems with the conjugate solvers, that occur especially in high dimension}. This is why we have favored the use of ELU~\citep{clevert2015fast} activations in the $u$ parameterization rather than Relu activations.

\subsection{Computation of $u_{\theta}^*$}
The use of a conjugate solver is necessary to compute the loss functions of $V_{\phi}$, $M_{\xi}$, $I_{\psi}$ and to estimate $M_{\theta}$. In all cases, the objective is to estimate the gradient of $u_{\theta}$'s conjugate at a given point $y$: $\nabla (u_{\theta})^*(y)$. For a given experiment, that justifies the use of the same conjugate solver parameters for these different applications. We relied on ADAM solver for the computation of the convex conjugate as it runs faster than LBFGS on our examples and use \citeauthor{amos2023amortizing} implementation. The two hyperparameters that remain to be set are the maximum number of iterations given to the solver to converge and the tolerance factor at which the norm of the gradient is considered small enough for the solver to have converged. These two hyperparameters are strongly dependent on the dimension of the problem as well as on the function $u_{\theta}$ and, therefore, on the distributions $\rho$ and $F_{\#} \rho$. To amortize the number of iterations required for the solver to converge, it is always initialized with the prediction of the $V_{\phi}$ network that is trained in conjunction with $u_{\theta}$. 
\subsection{Parameterize $V_{\phi}$} We use an MLP with 2 hidden layers of size $512$ and Relu activation functions to parameterize $V_{\phi}$ in all our experiments. 
\subsection{Parameterize $M_{\xi}$} The measure-preserving map $M$ is parameterized by a neural network only when the vector field under study is available through samples only. This is the case in the topography examples where $M_{\xi}$ is parameterized by an MLP with 2 hidden layers of size $512$ and Relu activation functions.
\subsection{Parameterize $X_{\psi}$} 
The learned part of the drift $X_{\psi}$ is parameterized using an MLP, and we use Silu activation functions which is the classic choice for parameterizing the drift $X_{\psi}$. The same hyperparameters have been used for the Chamonix, London, and Cyprus cases.

\begin{figure}
\centering
\begin{tabular}{lll}
\toprule
model & hyperparameter & value \\
\midrule
\multirow[t]{10}{*}{$u_{\theta}$} & activation function & elu \\
 & architecture & [64, 64, 64, 64] \\
 & b1 & 0.5 \\
 & b2 & 0.5 \\
 & scheduler & cosine decay \\
 & initial learning rate & 0.001 \\
 & $\alpha$ & 0.1 \\
 & scheduler steps & 50000 \\
 & steps & 50000 \\
\cmidrule{1-3}
\multirow[t]{9}{*}{$I_{\psi}$} & activation function & silu \\
 & architecture & [256, 256, 256] \\
 & scheduler & cosine decay \\
 & initial learning rate & 0.001 \\
 & $\alpha$ & 0.01 \\
 & scheduler steps & 50000 \\
 & steps & 50000 \\ 
 & $\sigma$ & 0.1 \\
\cmidrule{1-3}
\multirow[t]{9}{*}{$V_{\phi}$} & activation function & relu \\
 & architecture & [512, 512] \\
 & b1 & 0.9 \\
 & b2 & 0.999 \\
 & scheduler & cosine decay \\
 & initial learning rate & 0.0005 \\
 & $\alpha$ & 0.01 \\
 & scheduler steps & 50000 \\
 & steps & 50000 \\
\cmidrule{1-3}
\multirow[t]{4}{*}{conjugate solver} & name & Adam \\
 & max iteration & 200 \\
 & gtol & 0.001 \\
\cmidrule{1-3}
\bottomrule
\end{tabular}
\caption{Hyperparameters used for the topography experiments (the same hyperparameters have been used for Chamonix, London, and Cyprus).}
\end{figure}

\begin{figure}
\centering
\begin{tabular}{lll}
\toprule
model & hyperparameter & value \\
\midrule
\multirow[t]{10}{*}{$u_{\theta}$} & activation function & elu \\
 & architecture & [128, 128, 128, 128] \\
 & b1 & 0.5 \\
 & b2 & 0.5 \\
 & scheduler & cosine decay \\
 & initial learning rate & 0.001\\
 & $\alpha$ & 0.01 \\
 & scheduler steps & 10000 \\
 & steps & 10000 \\
\cmidrule{1-3}
\multirow[t]{9}{*}{$I_{\psi}$} & activation function & silu \\
 & architecture & [512, 512] \\
 & scheduler & cosine decay \\
 & initial learning rate & 0.0005 \\
 & $\alpha$ & 0.01 \\
 & scheduler steps & 50000 \\
 & steps & 50000 \\
 & $\sigma$ & 1.0 \\
\cmidrule{1-3}
\multirow[t]{9}{*}{$V_{\phi}$} & activation function & relu \\
 & architecture & [512, 512] \\
 & b1 & 0.9 \\
 & b2 & 0.999 \\
 & scheduler & cosine decay \\
 & initial learning rate & 0.0005 \\
 & $\alpha$ & 0.01 \\
 & scheduler steps & 10000 \\
 & steps & 10000 \\
\cmidrule{1-3}
\multirow[t]{4}{*}{conjugate solver} & name & Adam \\
 & max iterations & 700 \\
 & gtol & 0.1 \\
\cmidrule{1-3}
\bottomrule
\end{tabular}
\caption{Hyperparameters used for experiment 6.3.}
\end{figure}

\begin{figure}
\centering
\begin{tabular}{lll}
\toprule
model & hyperparameter & value \\
\midrule
\multirow[t]{10}{*}{$u_{\theta}$} & activation function & elu \\
 & architecture & [128, 128, 128, 128] \\
 & b1 & 0.5 \\
 & b2 & 0.5 \\
 & scheduler & cosine decay \\
 & initial learning rate & 0.0001 \\
 & $\alpha$ & 0.1 \\
 & scheduler steps & 30000 \\
\cmidrule{1-3}
\multirow[t]{8}{*}{$I_{\psi}$} & activation function & silu \\
 & architecture & [512, 512] \\
 & scheduler & constant learning rate \\
 & learning rate & 0.0005 \\
 & $\sigma$ & 0.1 \\
\cmidrule{1-3}
\multirow[t]{9}{*}{$V_{\phi}$} & activation function & relu \\
 & architecture & [512, 512] \\
 & b1 & 0.9 \\
 & b2 & 0.999 \\
 & scheduler & constant learning rate \\
 & learning rate & 0.0005 \\
\cmidrule{1-3}
\multirow[t]{4}{*}{conjugate solver} & name & Adam \\
 & max iteration & 1000 \\
 & gtol & 0.001 \\
\cmidrule{1-3}
\multirow[t]{9}{*}{particles} & steps & 60000 \\
 & particules & 1024 \\
 & $\gamma$ (step size for the gradient descent) & 0.1 \\
\cmidrule{1-3}
\bottomrule
\end{tabular}
\caption{Hyperparameters used for experiment 6.4.}
\end{figure}

\begin{figure}
\centering
\begin{tabular}{lll}
\toprule
model & hyperparameter & value \\
\midrule
\multirow[t]{10}{*}{$u_{\theta}$} & activation function & elu \\
 & architecture & [128, 128, 128, 128] \\
 & b1 & 0.5 \\
 & b2 & 0.5 \\
 & scheduler & cosine decay \\
 & initial learning rate & 0.0001 \\
 & $\alpha$ & 0.1 \\
 & scheduler steps & 30000 \\
\cmidrule{1-3}
\multirow[t]{8}{*}{$I_{\psi}$} & activation function & silu \\
 & architecture & [512, 512] \\
 & scheduler & constant learning rate \\
 & learning rate & 0.0005 \\
 & $\sigma$ & 0.1 \\
\cmidrule{1-3}
\multirow[t]{9}{*}{$V_{\phi}$} & activation function & relu \\
 & architecture & [512, 512] \\
 & b1 & 0.9 \\
 & b2 & 0.999 \\
 & scheduler & constant learning rate \\
 & learning rate & 0.0005 \\
\cmidrule{1-3}
\multirow[t]{4}{*}{conjugate solver} & name & Adam \\
 & max iteration & 1000 \\
 & gtol & 0.001 \\
\cmidrule{1-3}
\multirow[t]{9}{*}{particles} & steps & 60000 \\
 & LMC multiplicative coefficient in front of $\nabla f$ & 1000 \\
 & LMC multiplicative coefficient in front of $\nabla u$ & 1000\\
 & particules & 1024 \\
 & warming steps & 30000 \\
 & $N$ & 200 \\
 & $\tau_f$ (step size for LMC steps on $f$) & 0.0001 \\
 & $\tau_u$ (step size for LMC steps on $u$) & 0.0001 \\
\cmidrule{1-3}
\bottomrule
\end{tabular}
\caption{Hyperparameters used for experiment 6.5.}
\end{figure}

\end{document}